\documentclass[10pt,journal,compsoc]{IEEEtran}
\usepackage{graphicx}
\usepackage{multirow}
\usepackage{array}
\usepackage{makecell}
\usepackage{booktabs}
\usepackage{caption}
\usepackage[normalem]{ulem}
\usepackage{color}
\usepackage{xcolor}
\usepackage{url}

\ifCLASSOPTIONcompsoc
  \usepackage[nocompress]{cite}
\else
  \usepackage{cite}
\fi
\ifCLASSINFOpdf
\else
\fi
\begin{document}
\title{VERAM: View-Enhanced Recurrent Attention Model for 3D Shape Classification}

\author{Songle~Chen,~
        Lintao~Zheng,~
        Yan~Zhang,~
        Zhixin~Sun~
        and~~Kai~Xu* % <-this % stops a space
\IEEEcompsocitemizethanks{\IEEEcompsocthanksitem Corresponding author: Kai Xu (kevin.kai.xu@gmail.com).
\IEEEcompsocthanksitem Songle Chen is with Jiangsu High Technology Research Key Laboratory for Wireless Sensor Networks, Nanjing University of Posts and Telecommunications. Email: chensongle@njupt.edu.cn.
% \\ is fragile and will error, could use \hfil\break
\IEEEcompsocthanksitem Lintao Zheng is with School of Computer, National University of Defense
Technology. Email: lintaozheng1991@gmail.com.
\IEEEcompsocthanksitem Yan~Zhang is with Department of Computer Science and Technology, Nanjing University. Email: zhangyannju@nju.edu.cn.
\IEEEcompsocthanksitem Zhixin~Sun is with Jiangsu High Technology Research Key Laboratory for Wireless Sensor Networks, Nanjing University of Posts and Telecommunications. Email: sunzx@njupt.edu.cn.
\IEEEcompsocthanksitem Kai Xu is with School of Computer, National University of Defense Technology. Email: kevin.kai.xu@gmail.com.
}% <-this % stops an unwanted space
\thanks{Manuscript received Sep. 27, 2017; revised May 26, 2018.}}


\markboth{Journal of IEEE Transactions on Visualization and Computer Graphics,~Vol.~$\times\times$, No.~$\times\times$, $\times\times$~2017}%
{Shell \MakeLowercase{\textit{et al.}}: Bare Demo of IEEEtran.cls for Computer Society Journals}




\IEEEtitleabstractindextext{%
{\leftskip=0pt \rightskip=0pt plus 0cm

\begin{abstract}
Multi-view deep neural network is perhaps the most successful approach in 3D shape classification.
However, the fusion of multi-view features based on max or average pooling lacks a view selection mechanism,
limiting its application in, e.g., multi-view active object recognition by a robot.
This paper presents VERAM, a view-enhanced recurrent attention model capable of actively selecting
a sequence of views for highly accurate 3D shape classification.
VERAM addresses an important issue commonly found in existing attention-based models,
i.e., the unbalanced training of the subnetworks corresponding to next view estimation and
shape classification. The classification subnetwork is easily overfitted while the
view estimation one is usually poorly trained, leading to a suboptimal classification performance.
This is surmounted by three essential \emph{view-enhancement strategies}:
1) enhancing the information flow of gradient backpropagation for the view estimation subnetwork, 2) devising a highly informative reward function for the reinforcement training of view estimation and 3) formulating a novel loss function that explicitly circumvents view duplication. {\color{black} Taking grayscale image as input and AlexNet as CNN architecture, VERAM with $9$ views achieves instance-level and class-level accuracy of $95.5\%$ and $95.3\%$ on ModelNet10,  $93.7\%$ and $92.1\%$ on ModelNet40, both are the state-of-the-art performance under the same number of views.} %{\color{red}\sout{VERAM achieves an average classification accuracy of $96.1\%$ ($7$ views) on ModelNet10 and$91.5\%$ ($6$ views) on Modelnet40, both representing the state-of-the-art performance.}}
\end{abstract}
}

}

\maketitle

\IEEEdisplaynontitleabstractindextext

% \fi
\IEEEpeerreviewmaketitle

\IEEEraisesectionheading{\section{Introduction}\label{sec:introduction}}

\IEEEPARstart{3}{D} shape classification is a fundamental problem in the field of computer graphics and computer vision. It finds applications from traditional computer aided design and medical imaging to cutting-edge mixed reality and robot navigation. The challenge of 3D shape classification stems from
the difficulty of characterizing 3D surface geometry, the variety of 3D transformation and deformation,
and the imperfection of geometry and/or topology, etc.
Hundreds of hand-crafted 3D shape descriptors have been proposed, either from 2D rendered views\cite{chendy2003a,shihjl2007a,cyrmc2001a,ulrichm2012a,papadakisp2010a} or directly on 3D models \cite{ankerstm1999a, kazhdanm2003a,knoppj2010a,bronsteinam2011a}. They are, however, often carefully designed to characterize only one or a few aspects of 3D shapes, making them hard to generalize well.
%Their description and discrimination capacity cannot
%adapt to the diversity of numerous 3D shapes in different categories,
%to achieve satisfactory classification performance.

Recently, inspired by the advances in image classification using convolutional neural networks (CNN)~\cite{krizhevskya2012a,Simonyank2014a}, multi-view CNN (MVCNN) was presented for 3D shape classification~\cite{suh2015a,qicr2016a,wangc2017a}.
%{ \color{red}\sout{\mbox{\cite{shib2015a}}}}
By leveraging massive image databases such as ImageNet \cite{dengj2009a} to pre-train the CNN and learn image descriptors for general vision tasks, MVCNN has significantly advanced the
state-of-the-art of 3D shape classification.
The method renders a 3D shape to RGB or depth images from different viewpoints, uses {\color{black}advanced CNN architecture}
%{\color{red}\sout{pre-trained CNN}}
to extract features for each view, and then aggregates the multi-view features to form the final feature representation based on max or average pooling.
Albeit being simple yet effective, its best performance is achieved only when \emph{all} views are used.
In some practical scenarios, such as robot-operated active recognition, it is desirable to achieve object recognition with as-few-as-possible views, to minimize the robot movement cost.

%it is more suitable for the scenarios where the viewpoints of the rendered images of the 3D shape are unknown. However, when the viewpoint locations are available, the strategy of view pooling across \emph{all} views has limitations. %{\color{red}\sout{a limitation.}}
%{\color{red}\sout{It lacks a view selection mechanism and thus cannot exploit the complementary discriminative capability of different views, for the purpose of efficient classification with an as-few-as-possible set of views.}}
%High computational cost is involved both at training and testing,since the amount of computation scales is coupled with the number of views.}

Our key observation is that human is able to recognize a 3D shape without processing all views.
Given the first view observation of a 3D shape, a human tends to first form hypotheses about which categories the shape may fall in, and then switch to the next viewpoint purposely, by moving himself around the shape or rotating the shape, to quick narrow down the uncertainty and refine hypotheses. This process is repeated for a few times until sufficient evidence are collected to minimize the uncertainty.
There are two most prominent characters of the above procedure.
First, the informative view is quite sparsely selected, far from being exhaustive.
Second, both the next view selection and the predication making
are deduced from the combined information from the previous observations over time.

\captionsetup{skip=4pt}
\begin{figure*}
\centering
\includegraphics[width=6.5in]{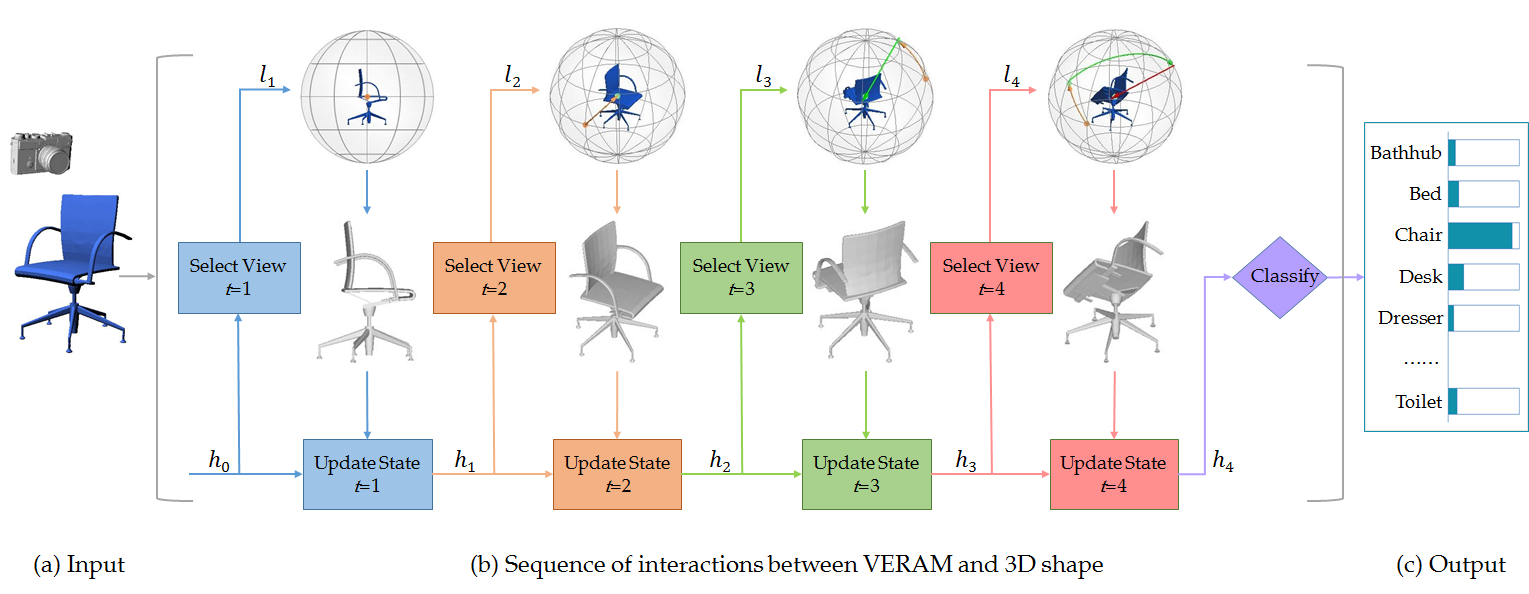}
\caption{VERAM is a view-enhanced recurrent attention model capable of adaptively selecting a sequence of views to classify 3D shapes. (a) The input is an unknown 3D shape to be rendered with a virtual camera. (b) The sequence of interactions between VERAM and the 3D shape. At each time step, VERAM actively selects the next view $l$ for rendering according to current internal state $h$, which is then updated by the new observation. (c) Based on the aggregated information over time steps, VERAM makes a classification and outputs the category probabilities of the input 3D shape.}
\label{fig:model}% give the label a meaningful name, for example, I use 'model'
\end{figure*}

Following the above intuition and drawing inspiration from visual attention model based on recurrent neural networks (RNN) \cite{baj2014a,mnihv2014a}, we present \emph{View-Enhanced Recurrent Attention Model (VERAM)} capable of automatically selecting an as short as possible sequence of views to classify 3D shapes.
As shown in Fig.~\ref{fig:model}, our model is formulated as a goal-directed agent interacting with a 3D shape by using a camera sensor (Fig.~\ref{fig:model}(a)).
At each time step, the agent actively selects the next viewpoint $l$ for the camera sensor, according to the current internal state $h$. Next, the camera sensor captures the 3D shape from the new viewpoint to obtain a 2D image
and pass it to the agent.
The agent then extracts features for the new input image and updates its internal state.
Such process is repeated for a few steps ($4$ time steps in Fig.~\ref{fig:model}(b)). Finally, based on the aggregated information over time steps, the agent emits the predication as the probabilities of shape categories (Fig.~\ref{fig:model}(c)).

On the technical side, our method has two advantages over MVCNN.
First, when pooling across all views, MVCNN discards the location information of each view. However, recent works reveal that viewpoint location plays an import role in enhancing the performance of the task of 3D shape classification \cite{johnse2016a,kanezakia2018a}. Second, processing all views involves high computational cost both at training and testing, although not every view is essential to recognition.

There have been a number of works using RNN-based visual attention model for image classification \cite{baj2014a,mnihv2014a,sermanetp2014a,semeniutas2016a}, image captioning \cite{xuk2015a}, and even for 3D object retrieval \cite{xuk2016a}.
However, a common issue with these RNN-based attention models is the unbalanced training of the
subnetworks corresponding to next view estimation and shape classification.
The classification subnetwork is usually easier to train than the view estimation one.
Consequently, the model training can be easily trapped in a local minima:
the classification subnetwork is overfitted while the view estimation one poorly trained,
which in turn leads to degraded classification accuracy.
In this work,
we propose three key technical contributions with VERAM, aiming
to enhance the view learning and achieve a significant performance boost to multi-view 3D shape classification.
%Our method outperforms not only the standard MVCNN but also the state-of-the-art attention-based models~\cite{xuk2015a} in 3D shape classification.
%thus providing a compact framework of supporting learning with location confidence and distribution integrally.
%The distinctive characteristics of VERAM include:
\begin{itemize}
  \item We introduce three schemes to improve the information flow of gradient backpropagation from the
   view estimation subnetwork to the hidden units, achieving a balance with the training of the classification subnetwork. This overcomes the issue that the estimated views may get stuck at the boundary in the view parameterization space, which is usually encountered in previous works~\cite{mnihv2014a}.
  \item During the reinforcement learning of our attention model, we integrate the classification confidence of the current view into the gradient computation of the reward against the view. This leads to a highly efficient guidance to the next view estimation.
  \item In VERAM, a novel loss function is proposed with a regularization term that enforces the estimated view to be distant to any of the previous ones so as to avoid view duplication.
\end{itemize}
	
The hybrid architecture of VERAM is trained for the subnetworks of shape classification and view estimation
jointly, with the former using SGD \cite{rumelhartde1986a} and the latter taking REINFORCE \cite{williams1992a}.
We empirically evaluate our model on the ModelNet benchmark \cite{wuz2015a}.
Taking rendered gray-scale image as input and AlexNet as CNN architecture, without applying any data augmentation or network ensemble strategy, VERAM with $9$ views achieves average instance-level and class-level accuracy of $95.5\%$ and $95.3\%$ on ModelNet10,  $93.7\%$ and $92.1\%$ on ModelNet40.
 %{\color{red}\sout{$91.5\%$ (with $6$ views) on Modelnet40, both representing the state-of-the-art performance.}}
 %Moreover, VERAM achieves recognition with only a limited number of views and runs in about $0.1$ second per shape on a TITAN X GPU.
 The high accuracy and efficiency make our model scalable to large datasets and applicable to many online applications. %{\color{blue}The \emph{source code} as well as the trained models of VERAM will be released at XXX.}

%{\color{red}\sout{The \emph{source code} as well as the trained state-of-the-art models of VERAM are now available at 180.209.64.122 for reviewing, and will be released on publicly upon publishing.}

\section{Related work}
\emph{\textbf{3D shape classification via hand-craft descriptors.}} There is a long history of work in 3D shape analysis and a large variety of hand-craft shape descriptors have been presented. The representative view-based descriptors cover Light Field descriptor \cite{chendy2003a}, Elevation descriptor \cite{shihjl2007a}, Aspect Graph based descriptor \cite{cyrmc2001a, ulrichm2012a} and DFT/DTW panoramic descriptor \cite{papadakisp2010a}, etc. Popular shape descriptors include Shape Histogram descriptor \cite{ankerstm1999a}, Spherical Harmonic descriptor \cite{kazhdanm2003a}, 3D SURF \cite{knoppj2010a}, Heat Kernel Signatures \cite{bronsteinam2011a}, etc. These descriptors are largely hand-engineered and usually do not have enough generalization ability to adapt to the diversity of numerous 3D shapes in different categories. As a result, their performance has an obvious gap compared with the current dominant methods based on deep learning technology, which have achieved state-of-the-art performance in many tasks of computer graphics and computer vision. Our method falls into the deep learning class.

\noindent\emph{\textbf{3D shape classification via deep CNN.}} A number of methods based on deep CNN have achieved state-of-the-art performance in 3D shape classification on public benchmarks. There are two categories. Shape-based methods \cite{qicr2016a,wuz2015a,maturanad2015a,brocka2016a,liy2016a,qicr2016b,qicr2017a,wangps2017a},{\color{black}\mbox{\cite{lij2018a}}} perform convolutions with 3D filters on the voxels or point clouds in continuous 3D space, and the volumetric representation makes them have the ability of exploiting complete structure information. View-based methods \cite{suh2015a,Xie2015,qicr2016a,wangc2017a,johnse2016a,kanezakia2018a,shib2015a,Sfikask2017a,Sfikask2017b} first render the 3D shape into 2D images from different viewpoints, and then apply 2D filters to carry out convolution for each view.

Compared with shape-based methods, the advantage of view-based methods is that the massive image databases can be used to pre-train the deep neural network and advanced network architectures succeeded in image recognition tasks can be employed. Partly for these reasons, to data, view-based methods shows better or comparable performance to shape-based methods. Moreover, convolution on 2D images is more efficient than on 3D volumetric space. View-based methods naturally need to fuse clues from different 2D views, and max or average pooling is the most common strategy to perform the task \cite{suh2015a,qicr2016a}, \cite{wangc2017a}
%{\color{red}\sout{\mbox{\cite{shib2015a}}}}
, which lacks a view selection mechanism. Our method is view-based and also employs deep CNN to extract the descriptors for the rendered views. However, we adopt RNN-based visual attention model to learn attention policy of adaptively selecting a few number discriminative views, which is more effective and efficient.

\noindent\emph{\textbf{3D shape recognition via active view selection.}} Our method fits into the realm of active recognition. Indeed, active recognition through next view planning has been studied for quite a long time in computer vision \cite{royds2004a}. For 3D object recognition and pose estimation, next-best-view selection based on information rich model was proposed in \cite{wuk2015a}. In each step, the next-best-view is selected as the voxels of which has the highest number of matches that have not been detected before. In 3DShapeNets framework \cite{wuz2015a}, next-best-view for 2.5D recognition is selected according to which can maximize the mutual information to reduce the potential uncertainty. In contrast to these local optimum next-best-view selection methods, an approximately global optimum approach was proposed by using undirected graph search \cite{johnse2016a}. However, the next-best-view selection is isolated from the neural network,  and it is not a completely global optimum method. By contrast, the next-best-view predication of our method is embedded in deep RNN and is a global optimum approach. Recently, MV-RNN \cite{xuk2016a} combines RNN-based visual attention model with MVCNN \cite{suh2015a} for 3D object retrieval. The view confidence and view location constrains are implicitly handled in the layer of feature representation. In contrast, VERAM does not combine MVCNN, and explicitly integrates view confidence and view location constrains into reward gradient and classification loss.

\noindent\emph{\textbf{Visual attention model based on RNN.}} We draw inspiration from recent approaches that used RNN-based visual attention model to learn task-specific policies in various applications. The vision tasks include image classification \cite{baj2014a,mnihv2014a,sermanetp2014a,semeniutas2016a} , image caption generation \cite{xuk2015a}, action with its boundary detection \cite{yeungs2016a}, and 3D object retrieval \cite{xuk2016a}. The attention model is also used in non-visual task, such as learning policies for a Neural Turing machine \cite{zarembaw2015a}. Our method builds on these directions and learns policies addressing the task for 3D shape classification. It extends the RNN-based visual attention models to be more robust to the common issue of the unbalanced training of the subnetworks, and provides a paradigm of how RNN-based attention model to support learning with view confidence and view location constrains integrally.

\section{Method}
\begin{figure}[h]
\centering
\includegraphics[width = .50\textwidth]{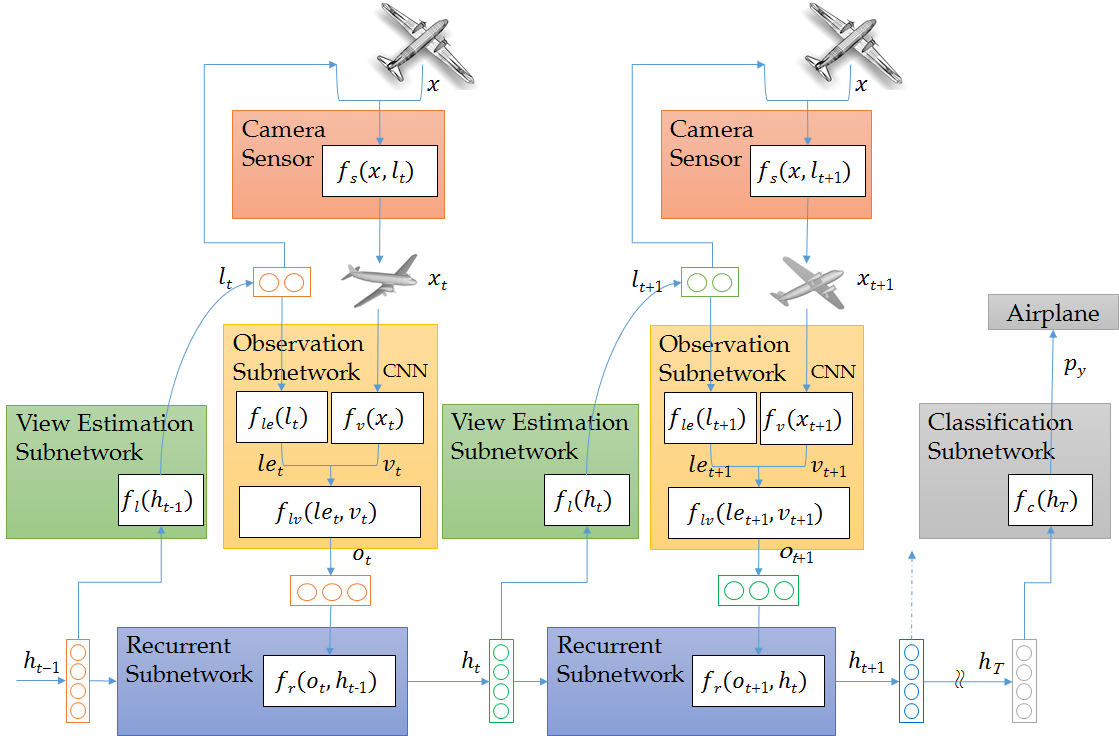}
\caption{Architecture of VERAM. Its sub-components include a virtual camera sensor, observation subnetwork, recurrent subnetwork, view estimation subnetwork and classification subnetwork. }
\label{fig:architecture}
\end{figure}

The proposed VERAM is a RNN-based visual attention model for 3D shape classification, and is formulated as a goal-directed agent interacting with a 3D shape. A graphical representation of VERAM is shown in Fig.~\ref{fig:architecture}.

\setlength{\intextsep}{5pt plus 1pt minus 1pt}

The model sequentially processes a 3D shape $x$ within $T$ time steps. At each step $t$, based on the internal state $h_{t-1}$, the model actively selects a viewpoint $l_t$ and obtains an observation $x_t$ of 2D image rendered by the camera sensor from $l_t$, and then, the model uses $x_t$ with $l_t$ to updates its internal state. This process is repeated until the predication is emitted at step $T$. The architecture of the model will be described in subsection 3.1 and later in subsection 3.2, we explain how to use a combination of SGD and REINFORCE to train the model in end-to-end fashion.

\subsection{Architecture}
 The architecture of VERAM can be broken down into a number of sub-components including camera sensor, observation subnetwork, recurrent subnetwork, view estimation subnetwork and classification subnetwork. Each component maps the input into a matrix or vector output.

%\IEEEpubidadjcol

\subsubsection{Camera sensor}
\begin{figure}[h]
\centering
\includegraphics[width = .45\textwidth]{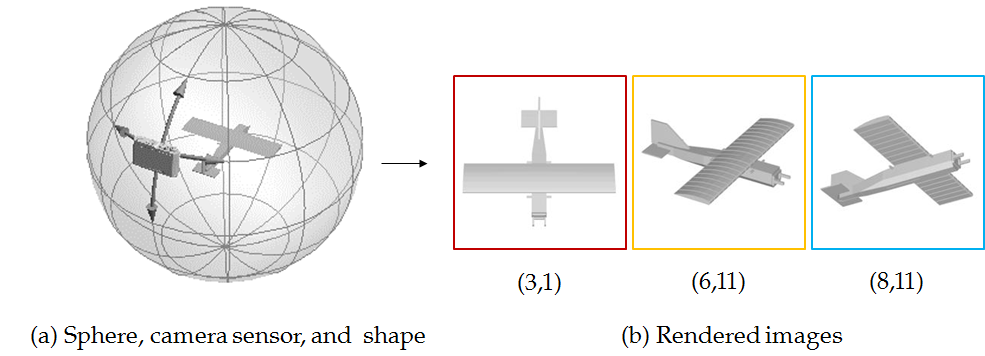}
\caption{3D shape located in the center of the viewing sphere is rendered into 2D images by the camera sensor. }
\label{fig:viewsphere}
\end{figure}

As shown in Fig.~\ref{fig:architecture}, the input to camera sensor is a 3D shape $x$ and viewpoint location $l_t$, the output of the camera sensor is the 2D image $x_t$ of the rendered view. 3D shape $x$ represented as polygon mesh is located at the center of the viewing sphere, as shown in Fig.~\ref{fig:viewsphere} (a). The camera sensor can move on the surface of the viewing sphere and its location is indicated by the latitude and longitude. The camera sensor always points towards the centroid of the shape, and its upright vector is the tangent line of the latitude along clockwise direction. Phong reflection model is used to render the 3D shape into 2D images ($224\times224$). Under a perspective projection, the pixel color is determined by the reflected intensity of the polygon vertices. Example rendered images are shown in Fig.~\ref{fig:viewsphere} (b).%When the context is clear, we refer to a view and its corresponding rendered image interchangeably.

\begin{figure}[h]
\centering
\includegraphics[width = .4\textwidth]{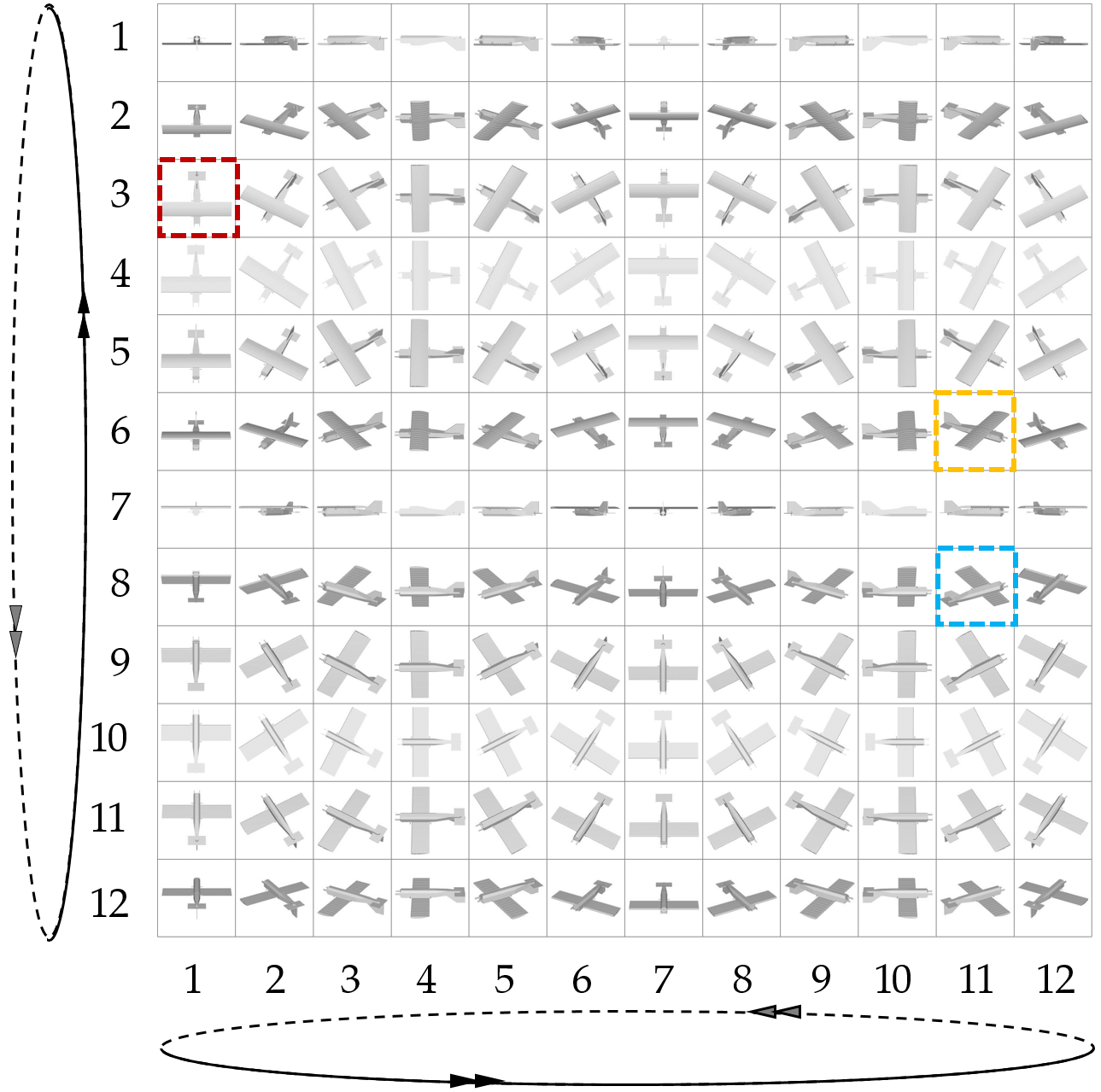}
\caption{Sample space of the viewpoint locations on viewing sphere is mapped to 12$\times$12 grid for camera sensor, which defines the view parameterization space for VERAM.}
\label{fig:viewpointgrid}
\end{figure}

Data preparation is necessary for efficient training, and we sample discrete views at every $30$ degrees both in latitude and longitude. If all views of a shape are rendered, they can be arranged in $12\times12$ grid, as shown in Fig.~\ref{fig:viewpointgrid}. The ordinal number $1$ to $12$ in the grid along vertical and horizontal direction is corresponding to the latitude and longitude of the sampled viewpoint location, which defines the view parameterization space for VERAM.

There are two advantages of our sample strategy. First, the entire sphere is uniformly sampled both on latitude and longitude. Second, along the horizontal or vertical direction, the $12$ images are consecutively connected and form a circle. As a result, it provides a contiguous space for the agent to deploy the camera sensor in an absolute or relative coordinate. In the following sections, the absolute viewpoint location in the view parameterization space is represented as $l = (r, c)$, and both $r$, $c$ are in the range of $[1, 12]$. The magnified images of $(3, 1)$, $(6, 11)$ and $(8, 11)$ in Fig.~\ref{fig:viewpointgrid} are shown in Fig.~\ref{fig:viewsphere} (b). The function of the camera sensor can be formulated as
\begin{equation}
x_t=f_s(x,l_t).
\label{eq:abc}
\end{equation}

\subsubsection{Observation subnetwork}
The job of the observation subnetwork is to encode the information about both where the observation is taken as well as what has been seen. As Fig.~\ref{fig:architecture} illustrates, at each time step $t$, the observation subnetwork takes the rendered image $x_t$ and location tuple $l_t = (r_t, c_t)$ as input and output a vector $o_t$.

We use $v_t = f_v(x_t;\theta_v)$ to denote the output $v_t$ from function $f_v (.)$ that takes image $x_t$ as input and is parameterized by weights $\theta_v$. $f_v (.)$ typically maps with a sequence of convolutional, pooling, and fully connected layers, and the output of which is high level features $v_t$. Advanced network architectures succeeded in image recognition task can be used for $f_v(.)$, such as AlexNet \cite{krizhevskya2012a}, VGG-16 \cite{Simonyank2014a}, ResNet \cite{hek2016a}, etc. The location tuple $l_t$ is mapped into embedding $le_t$ by a fully connected layer $f_{le} (l_t;\theta_{le})$. In our practice, a fully connected layer is corresponding to a rectified linear unit $ReLU(Wx+b)$.

We concatenate the low bandwidth location $le_t$ with the high bandwidth view information $v_t$ by a fully connected layer $f_{lv}([le_t\ v_t];\theta_{lv})$ and output the final observation feature vector $o_t$. The observation subnetwork $f_o$ can be represented as
\begin{equation}
o_t = f_o (l_t, x_t;\theta_o) = f_{lv} ([f_{le} (l_t;\theta_{le})\ f_v (x_t;\theta_v)]; \theta_{lv}),
\label{eq:ot}
\end{equation}
where $\theta_o = [\theta_{le}, \theta_v, \theta_{lv}]$, corresponding to the whole parameters of the observation subnetwork.

\subsubsection{Recurrent subnetwork}
The agent maintains an internal state which encodes the agent's knowledge of the environment and summarizes information extracted from the history of past observations. It is instrumental to deciding how to act and where to deploy the camera sensor. In VERAM, this internal state is formed by the hidden units $h_t$ of the recurrent neural network and updated over each time step with the feature vector $o_t$ from the observation subnetwork, as shown in Fig.~\ref{fig:architecture}. The recurrent subnetwork is defined as
\begin{equation}
h_t = f_r (o_t, h_{t-1};\theta_r).
\label{eq:ht}
\end{equation}

linear mapping can be used for its efficiency, but Long-Short-Term Memory units {\color{black}(LSTM)} \cite{hochreiters1997a} has the ability to learn long-range dependencies and stable dynamics.
\subsubsection{View estimation subnetwork}
The view estimation subnetwork acts as a controller that directs attention based on the current internal state. In Fig.~\ref{fig:architecture}, the view estimation subnetwork takes the hidden units $h_{t-1}$ of recurrent subnetwork as input, and outputs $l_t$ to make a prediction on where to deploy the camera sensor to render the next view.
The view estimation subnetwork consists of $f_l (h_{t-1};\theta_l)$ and  $f_g (u_t;\theta_g)$. $f_l$ maps the hidden units $h_{t-1}$ into a two-dimensional coordinate tuple $u_t$, formally defined as
\begin{equation}
u_t = f_l (h_{t-1};\theta_l).
\label{eq:ut}
\end{equation}

$f_l$ is usually implemented by a $Linear$ layer followed by a specific transfer function. In the testing phase, $u_t$ is directly used as the next viewpoint location $l_t$. %for the camera sensor.
The procedure of obtaining the image $x_t$ from location $l_t$ is non-differentiable, so in the training phase, a stochastic module $f_g (u_t;\theta_g)$ needs to be used. It samples $l_t$ stochastically from a Gaussian distribution with a mean $u_t$ and a fixed variance $\delta$ for reinforcement learning, defined as
\begin{equation}
l_t = f_g (u_t;\theta_g) = N(u_t, \delta^2).
\label{eq:lt}
\end{equation}

The detail of view estimation subnetwork will be discussed in subsection 3.2.

\subsubsection{Classification subnetwork}
The classification subnetwork makes a classification and outputs the category probabilities $y$ of the input 3D shape $x$ based on the final internal state $h_T$, which integrates the information of the interaction history between the agent and input 3D shape. In VERAM, the classification subnetwork $f_c$ has a fully connected layer followed by $LogSoftMax$ output layer, namely
\begin{equation}
 P(y|x) = f_c (h_T;\theta_c).
\label{eq:py}
\end{equation}

\subsection{Learning}
Given the category label $y$ of shape $x$, we can formulate learning as a supervised classification problem. However, because the architecture of VERAM is hybrid, training it involves challenges of handling the non-differentiable component, of keeping balance between learning subnetworks, while struggling with the overfitting problem caused by millions of parameters. In this section, we will describe how VERAM combines SGD \cite{rumelhartde1986a} and REINFORCE \cite{williams1992a} to solve these problems.

\subsubsection{REINFORCE-based learning}
In subsection 3.1.1, the camera sensor is formulated as $f_s$ to map 3D shape $x$ with $l_t$ to 2D image $x_t$. However, $f_s$ is not a continuous function. As shown in Fig.~\ref{fig:boundary}, the gradient displayed as green arrow lines from observation subnetwork cannot back propagate to the view estimation subnetwork via the camera sensor. As a result, the view estimation subnetwork cannot be trained with standard back propagation.

REINFORCE \cite{williams1992a} is adopted to solve this problem. Given a space of action sequences $A$, $p(a)$ is a distribution over $a\in A$ and parameterized by $\theta$, we wish to learn network parameters $\theta$ that maximize the expected reward of action sequences. The gradient of the objective is
\begin{equation}
 \nabla J(\theta)= \sum\nolimits_{a\in A} p_{\theta}(a) \nabla log\;p_{\theta}(a)r(a).
\label{eq:jtheta}
\end{equation}

Here $r(a)$ is a reward assigned to each possible action sequence. Obviously, due to the high-dimensional space, this is a non-trivial optimization problem. REINFORCE addresses this by learning network parameters using Monte Carlo sampling. After running an agent's current policy $\pi _{\theta}$ in its environment and obtaining $K$ interaction sequences of length $T$, the approximation to the gradient equation is
\begin{equation}
 \nabla J(\theta)\approx 1/K\sum_{i=1}^{K}\sum_{t=1}^{T} \nabla log \;\pi _{\theta}(a_t^i|s_{1:t};\theta)(R_t^i-b_t).
\label{eq:jthetaapprox}
\end{equation}

Here, policy $\pi _{\theta}$ maps the history of past interactions with the environment $s_{1:t}$ to a distribution over actions for the current time step $t$. $R_t$ is the cumulative future reward from the current time step to time step $T$, $b_t$ is a baseline reward to reduce the variance of the gradient estimation. For a detail RNN-based REINFORCE, we refer to \cite{williams1992a} and \cite{wierstrad2007a}.

In our case, $a_t$ is to predict the next location $u_t$, $s_{1:t}$ is summarized in the state of the hidden units $h_{t-1}$. $R=1$ if the shape is classified correctly after $T$ steps and $0$ otherwise. Policy $\pi$ is a fully connected hidden layer that maps $h_{t-1}$ to $u_t$ as formally defined in formula (4), and $\theta$ is corresponding to $\theta_l$. Gaussian distribution is adopt to implement the reinforcement algorithm. Assuming $f$ is the density function of the Gaussian distribution $N$ as defined in formula (5), $l_t$ is the sampled location, the gradient of $R$ w.r.t. $u_t$ is
\begin{equation}
 \frac {dR}{du_t} = (R-b)\times \frac {d\;ln(f(l_t,u_t))}{du_t}.
\label{eq:dru}
\end{equation}

This gradient can easily back propagate to $\theta_l$, so $\nabla$$J(\theta)$ can then be calculated. The flow direction of this gradient is shown as red arrow lines in Fig~\ref{fig:boundary}. REINFORCE learns model parameters according to this approximate gradient. It increases the log-probability of an action with a larger than expected cumulative reward, and decreases the probability if the obtained cumulative reward is smaller.

\subsubsection{Enhancing the information flow of gradient}
\begin{figure}[h]
\centering
\includegraphics[width = .45\textwidth]{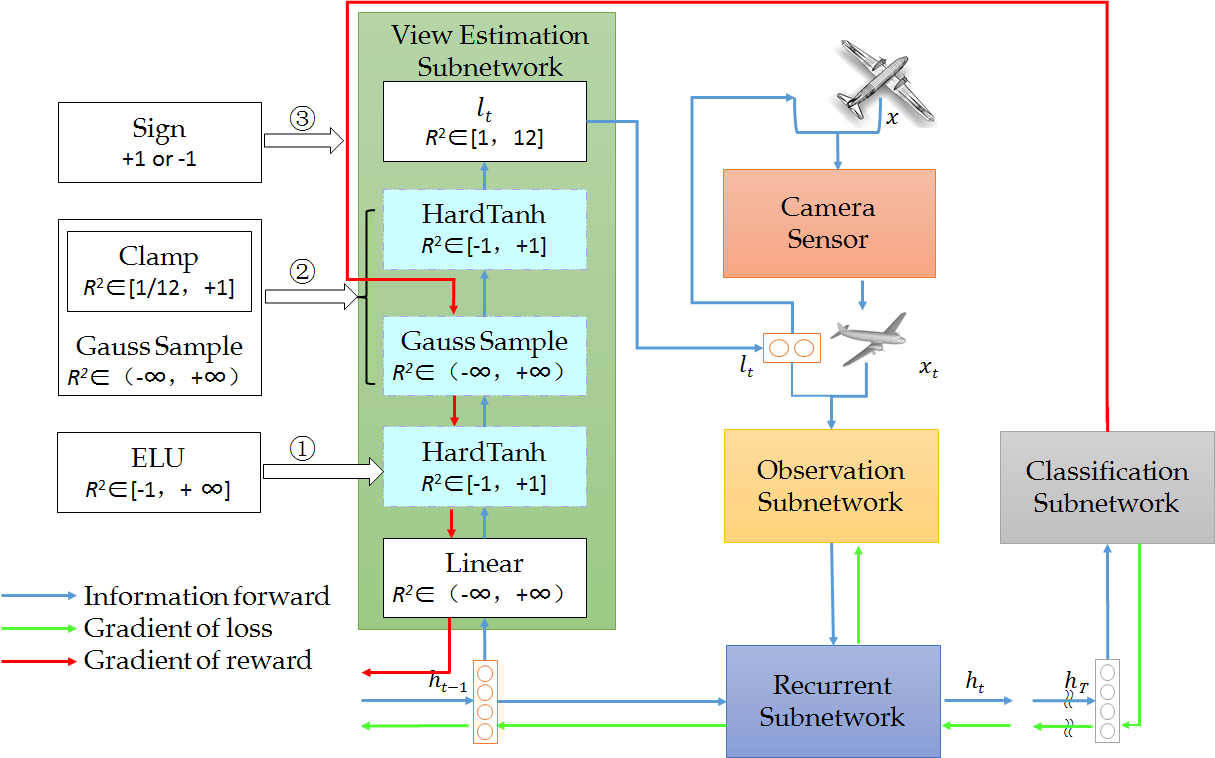}
\caption{Based on the visual attention model presented in \cite{mnihv2014a}, three schemes for enhancing information flow of gradient backpropagation for the view estimation subnetwork.}
\label{fig:boundary}
\end{figure}

As shown in Fig.~\ref{fig:boundary}, observation subnetwork, recurrent subnetwork, and classification subnetwork are with standard deterministic neural network connections, and can be trained directly by back propagating gradients from the classification loss, namely, by SGD. By contrast, view estimation subnetwork containing stochastic module needs to be trained using REINFORCE described in subsection 3.2.1.

Giving deep insight into the architecture as shown in Fig.~\ref{fig:architecture} and Fig.~\ref{fig:boundary}, we can find that there exists two ways starting from the hidden units $h_{t-1}$ and ending at classification subnetwork. Analogous to other RNN-based visual attention models, VERAM is essentially a parallel neural network, and suffers from the common issue of the unbalanced training of the subnetworks corresponding to next view estimation and shape classification. Specifically, as pointed in \cite{semeniutas2016a}, the full model does not have enough time to learn a good attention policy before the classification subnetwork overfits to the data. In our practice, we find regardless of different initial parameters and different 3D shapes, the estimated view locations $l_t$ tend to get stuck at the boundary of the view parameterization space, namely $1$ or $12$. Based on the visual attention model presented in \cite{mnihv2014a}, we propose three critical schemes to solve this issue, as shown in Fig.~\ref{fig:boundary}. It can be summarized as follows:

\indent 1) As mentioned in subsection 3.1.4, $f_l$ is implemented by a $Linear$ layer followed by a specific transfer function to force the estimated location $u_t$ into the target range. $HardTanh$ as well as $Sigmoid$ are commonly used for their continuous interval can be easily mapped to view  parameterization space. However, we find the problem of gradient disappearance caused by these transfer functions \cite{Hochreiters1998a} is very serious, which means in reinforcement learning, the gradient from reward cannot back propagate to hidden units $h$. VERAM adopts $Eula$ \cite{clevertda2015a} as this specific transfer function, which can effectively alleviate the vanishing gradient problem via the identity for positive values.

\indent 2) Similar to $f_l$, we also need a function for $f_g$ to force the output of $Gauss\ sample$ to fall into the view parameterization space. As shown in Fig.~\ref{fig:boundary}, the second $HardTanh$ operation is performed in \cite{mnihv2014a} for this purpose. However, the gradient of reward bypasses it and directly applies to $Gauss\ sample$ function. If the output $l_t$ of $Gauss\ sample$ is out of the range, the gradient calculated by formula (9) will be not right for $l_t$ is not the actual viewpoint location. To solve this problem, VERAM performs $Clamp$ in $Gauss\ sample$ itself to simulate the output of $Gauss\ sample$ is always in the range of view parameterization space.\\
\indent 3) In the backward process, formula (9) indicates that if the shape is misclassified, the learning process will encourage $u_t$ move away from $l_t$. In the boundary, this also needs to be refreshed with sign as
\begin{equation}
sign = \left\{
             \begin{array}{lcl}
             +1, {if \;l_t\; is\; not\; boundary} \\
             -1,  {if \;u_t<l_t, l_t=1/12,left \;boundary} \\
             -1,  {if\; u_t>l_t ,l_t=1, \quad right\; boundary}
             \end{array}.
        \right.
\label{eq:dru}
\end{equation}

Besides, if the shape is classified correctly, sign always sets to $+1$. Sign will multiplies to the result of formula (9). It means, in boundary, we still need to move $u_t$ to $l_t$ if $u_t$ is not in the range of [1/12, 1], although the shape is misclassified.

These three critical schemes enhance the information flow of gradient backpropagation from the view estimation subnetwork to the hidden units, and ensure each module be sequentially trained without break. As a result, it can effectively overcome the issue that the estimated views getting stuck at the boundary of the view parameterization space.

\subsubsection{ Learning with view confidence}
The schemes for enhancing the information flow of gradient provides a basis for keeping a balance between the subnetworks. However, the classification subnetwork is still easier to train than the view estimation one, making the learned attention policy be easily trapped in local optimization. In this subsection, we propose a method of learning with view confidence for REINFORCE to solve this problem.

As mentioned in subsection 3.1.2, advanced network architectures succeeded in image recognition tasks
can be used for $f_v(.)$ to extract features $v_t$ for image $x_t$. Besides, for all training shapes, we also extract the confidence $c_t$ of image $x_t$. Concretely, first, we extract features $v_t$ for each image $x_t$ of all views of the training shapes. Second, each image $x_t$ is also labeled with the same category $y_t$ of the 3D shape. By this means, for each 3D shape in the training set, we can obtain $144$ pairs of $(v_t, y_t)$. The collection of $(v_t, y_t)$ of all the 3D shapes in the training set will be taken as the input to the following simple network
\begin{equation}
 P(y_t|x_t) = LogSoftMax(Linear(v_t, \#categroies)).
\label{eq:dru}
\end{equation}

\begin{figure}[h]
\centering
\includegraphics[width = .35\textwidth]{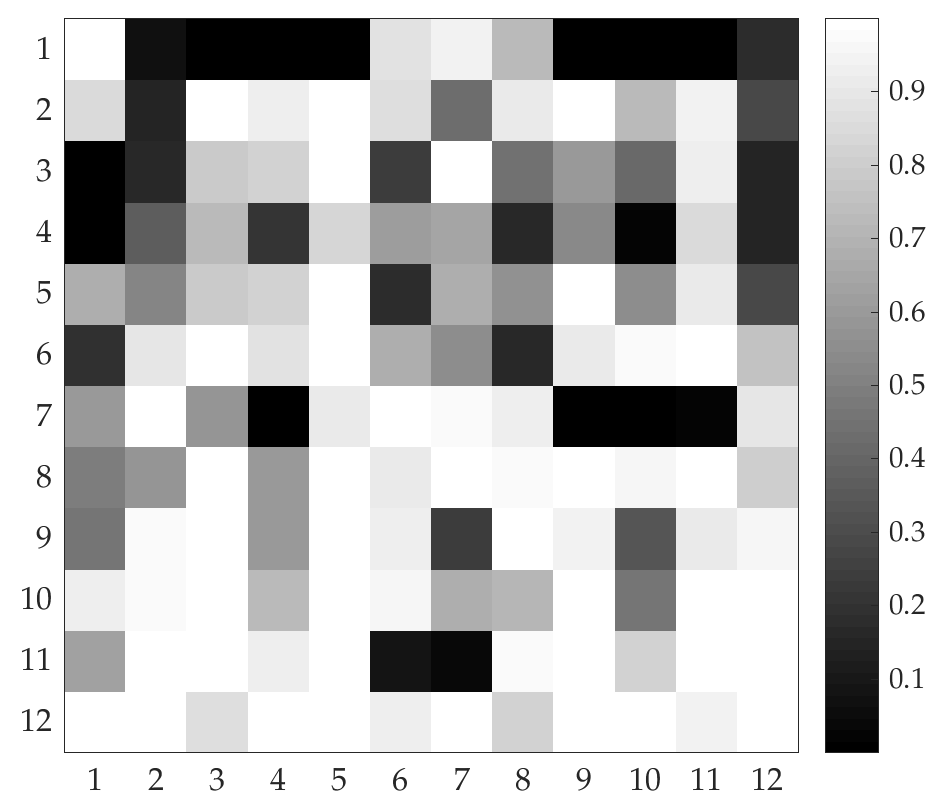}
\caption{Confidence, or category probability of each 2D image shown in Fig.~\ref{fig:viewpointgrid}.}
\label{fig:confidence}
\end{figure}

We use the negative log likelihood criterion to train this network. After convergence, the output probability of $LogSoftMax$ corresponding to its category is extracted as the confidence $c_t$ of image $x_t$. Fig.~\ref{fig:confidence} presents the confidence of each 2D image shown in Fig.~\ref{fig:viewpointgrid}. %{\color{red}\sout{The longer the bar}}
 {\color{black}The whiter the viewpoint}, the more confidence the image has. We can see the image in $(3, 1)$ almost has no confidence for its own category $airplane$. According to the output of $LogSoftMax$, the highest category probability of this image is $monitor$.

If a shape is classified correctly, according to formula (9), reinforcement learning will encourage all $u_{t+1}$ to move to $l_t$. On the contrary, it will encourage all $u_{t+1}$ to move away from $l_t$. Intuitively, if a shape is classified correctly, but in step $t$, the confidence $c_t$ of image $x_t$ located at $l_t$ is not high, we should weaken its effects of encouraging $u_{t+1}$ to move to $l_t$. On the other hand, if a shape is misclassified, but the confidence $c_t$ of image $x_t$ located at $l_t$ is high, we should weaken its effects of encouraging $u_{t+1}$ to move away from $l_t$. Based on this inspiration, combining formula (10), we refresh formula (9) as
\begin{equation}
 \frac {dR}{du_t} = sign \left\{
             \begin{array}{lcl}
             (R-b)\times  \displaystyle \frac {d\;ln(f(l_t,u_t))}{du_t}\times c_t\\
             (R-b)\times  \displaystyle \frac {d\;ln(f(l_t,u_t))}{du_t}\times (1-c_t) \rule{0ex}{2.0em}
             \end{array}
        \right..
\label{eq:dru}
\end{equation}

If a shape is classified correctly, the first line of formula (12) is used to calculate the gradient, otherwise, the second line of formula (12) is used. VERAM incorporates view confidence into reinforcement learning, which leads to a highly
efficient guidance to the next view estimation, and can effectively prevent the view estimation subnetwork from trapping into local optimization.

{\color{black}Note that confidence extraction is the preprocess procedure. The trained image classification network is only used to extract the confidence of image, and it will not be used anymore.}

%Note that: 1)Only the training set of 3D shape is used to train this image classification network and we stop training after convergence. For example, the instance-level accuracy of the training set of Modelnet40 converged at {75.34\%} by using the AlexNet FC6. 2)Confidence extraction is the preprocess procedure. The trained image classification network is only used to extract the confidence of image, and it will not be used anymore. 3)The extracted confidence is only used in the back propagation to adjust the gradient, so it don't need to extract the confidence of each image for the shape in the testing set.

\subsubsection{Learning with view location constrains}
In practice, we find the visited view of each time step may overlap. To solve this problem and make better use of the complementary capacities of different views. A novel regular term for view location constrains is supplemented to loss function and learned at the same time to make the distribution of the visited view locations much more reasonable and mutually complementary.

The regular term for the view location constrains is depend on the prior knowledge of the applications. For 3D shape classification, there are total $144$ view locations for the camera sensor, but VERAM only needs a few time steps $T$ to form the judgment, so the visited views at least should be separated from each other. This weak regular term is adopted by VERAM.

Particularly, after the agent has moved through $T$ steps, we add pairwise distance layer for each two view locations $l_i$ and $l_j$ the agent has visited. We train this part of the model with the loss of $HingeEmbeddingCriterion$
%\cite{collobertr2011a,leonardn2015a}
to force the visited views to separate from each other. The loss is formally defined as
\begin{equation}
Loss(l_i, l_j) = max(0, 1/12 - PairDistance(l_i, l_j)).
 \label{eq:dru}
\end{equation}

By integrating this loss function into the learning framework of recurrent attention model, VERAM can explicitly circumvents view duplication and the performance is further improved.

\section{Experiments and Discussions}
\color{black}{\subsection{Dataset, criteria and implementation details}}\color{black}
%\vspace{8pt}
\noindent\textbf{Dataset.}
We evaluate VERAM along with current state-of-the-arts on ModelNet10 and MondelNet40 \cite{wuz2015a}.
%{\color{red}\sout{, which are well annotated. }}
ModelNet10 contains $10$ categories with $4899$ %{\color{red}\sout{$4905$}}
unique 3D shapes, and ModelNet40 contains $40$ categories with $12311$ unique 3D shapes. The training set and testing set have been split on the website.
%{\color{red}\sout{Meanwhile, all 3D shapes have been aligned in their canonical orientation. We assume every shape keeping this orientation for fair comparison.}}

\noindent\textbf{Criteria.} We report classification accuracy
%{\color{red}\sout{on all shapes in the testing set }}
with two level criteria. Instance-level accuracy is the ratio of the number of shapes that are classified correctly to the number of the total shapes. {\color{black} Class-level accuracy is the average of instance-level accuracy among all categories. Class-level accuracy is more objective for there are big difference among different categories in the number of testing shapes, from 20 to 100, but instance-level is more intuitive for it directly reflects how many shapes are misclassified. }
%{\color{red}\sout{in the testing set. To calculate class-level accuracy, we first calculate the instance-level accuracy for each category separately, then average them among all categories.}}
 %{\color{red}\sout{Because}}
 %{\color{red}\sout{Class-level accuracy is more objective for }there are big difference among different categories in the number of testing shapes, from $20$ to $100$, we prefer to use the class-level accuracy if the data for comparison is available }}

\iffalse
\vspace{8pt}
\color{black}\noindent\textbf{Criteria. }\color{black}
We report classification accuracy
%{\color{red}\sout{on all shapes in the testing set }}
with two level criteria. Instance-level accuracy is the ratio of the number of shapes that are classified correctly to the number of the total shapes in the testing set. To calculate class-level accuracy, we first calculate the instance-level accuracy for each category separately, then average them among all categories. %{\color{red}\sout{Because}}
 {\color{black}Class-level accuracy is more objective for }there are big difference among different categories in the number of testing shapes, from $20$ to $100$,
 %{\color{red}\sout{we prefer to use the class-level accuracy if the data for comparison is available }}
 {\color{black}but instance-level is more intuitive for it directly reflects how many shapes are misclassified}.
\fi
\vspace{8pt}
\noindent\textbf{Implementation details.}
VERAM is implemented by Torch
%\cite{collobertr2011a,leonardn2015a}
on the platform with NVIDIA GeForce TITAN X. It needs about $1500$ epochs for training. The learning rate is set to $0.001$ in the first $600$ epochs, then decreases linearly to minimum $0.00001$ at epoch $1200$. Momentum is set to $0.9$.
%{\color{red}\sout{,and batch size is set to $20$}}
Based on grid search, the fixed variance $\delta$ of Gaussian distribution for REINFORCE is set to $0.11$.

{\color{black}
CNN architecture and recurrent subnetwork are two main components that affect the performance of VERAM. CNN is used to encode the rendered image. For CNN architecture, AlexNet \cite{krizhevskya2012a} and ResNet \cite{hek2016a} are used in our experiments. For recurrent subnetwork, linear mapping and LSTM \cite{hochreiters1997a} are adopted to update the hidden state in each time step. We will give the detail settings in each part of the evaluation.

Theoretically, the parameters of CNN in the observation subnetwork can be fine-tuned with the recurrent network at the same time. However, the training progress will be very slow because the forward and backward propagation need to proceed at each time step. For efficient training, the parameters of AlexNet and ResNet pre-trained on ImageNet are fixed and without further fine-tuning process on ModelNet. However, we can select more previous layer of CNN to counteract this influence.}

The \emph{source code} as well as the trained model of VERAM will be released at our project page:
\url{www.kevinkaixu.net/projects/veram.html}.

%CNN in the preprocess procedure of confidence extraction described in subsection 3.2.3 can be fine-tuned to get better performance of image classification. Meanwhile, CNN in the observation subnetwork described in subsection 3.1.2 also can be fine-tuned with the recurrent network at the same time. However, the training process is too slow for the forward and backward process are implemented at each time steps. For efficient training, the parameters of AlexNet and ResNet pre-trained on ImageNetare are fixed and without further fine-tune process on ModelNet.

%{\color{red}\sout{For each 2D image $x_t$, we extract features $v_t\in R^{2048}$ from layer $flatten0\_output$ of the ResNet152 pre-trained on dataset ImageNet 11K, which is available on MXNet \mbox{\cite{chent2015a}}.}}

{\color{black}\subsection{Comparison with state-of-the-arts}}
%\vspace{8pt}
%{\color{red}\sout{\noindent\textbf{Comparison with state-of-the-arts.}}}
{\color{black}\noindent In this subsection, we will compare the performance of VERAM with state-of-the-art deep neural-based methods, especially with the view-based methods. The performance of VERAM is achieved by taking AlexNet as CNN architecture for fair comparison. LSTM is adopted for recurrent subnetwork.

%{\color{red}\sout{\noindent\textbf{Comparison with state-of-the-arts.}}}
%{\color{black}\noindent In this subsection, we will compare the performance of VERAM with state-of-the-art deep neural-based methods on 3D shape classification. Specially, we will present the detail analysis for the view-based methods. The performance of VERAM is achieved by taking AlexNet as CNN architecture for fair comparison. LSTM is adopted for recurrent subnetwork.
\vspace{8pt}

\noindent\textbf{Comparison with deep neural-based methods.}}
%{\color{red}\sout{We compared the performance of VERAM on shape classification with that of state-of-the-art methods based on deep neural networks}}
{\color{black} The performance of VERAM is compared with state-of-the-art deep neural-based methods on 3D shape classification}, as summarized in table~\ref{table:cmpartsdeep}. These methods can be roughly grouped into two categories: shape-based and view-based. Shape-based methods cover 3DShapeNets \cite{wuz2015a}, VoxNet \cite{maturanad2015a}, SubVolSup \cite{qicr2016a}, AniProbing \cite{qicr2016a}, VRN {\color{black}\& VRN-Ensemble} \cite{brocka2016a}, FPNN \cite{liy2016a}, PointNet \cite{qicr2016b}, PointNet++ \cite{qicr2017a}, O-CNN \cite{wangps2017a} and \color{black}So-Net\cite{lij2018a}\color{black}. The view-based methods include MVCNN \cite{suh2015a}, DeepPano \cite{shib2015a}, {\color{black}PANORAMA-NN\cite{Sfikask2017a}, PANORAMA-ENN\cite{Sfikask2017b}}, MVCNN-Alex \cite{qicr2016a}, MVCNN-MultiRes \cite{qicr2016a},  Pairwise \cite{johnse2016a}, \color{black}DomSetClust \cite{wangc2017a},  RotationNet\cite{kanezakia2018a} \color{black}and {\color{black}the proposed} VERAM. Besides,  FusionNet \cite{hegdev2016a} exploits both volumetric representation and projective pixel representation.
%{\color{red}\sout{For fair comparison, similar to\mbox{ \cite{wangps2017a}}, we omit the results from the ensemble of models obtained by training the same deep neural network for many times, because it may involve some artifacts.}

%\sout{To alleviate the overfitting or to implement data augment, the strategies of multi-orientation, multi-view and multi-resolution are often adopted alone or in combinations. Although these strategies can lead to additional performance improvements, there would be a price to pay in terms of efficiency. In table~\ref{table:cmparts}, if the strategy is not adopted by the specific method, it is denoted as $\times$. Otherwise, the concrete number is marked for detail comparison. We use {-} to indicate the item is unavailable from the paper. The strategies of multi-orientation and multi-view are merged into the third column for they are the homogeneous type.}

%\vspace{2ex}
\newcommand{\tabincell}[2]{\begin{tabular}{@{}#1@{}}#2\end{tabular}}
\renewcommand\arraystretch{1.2}

\begin{table}[h]\footnotesize
\caption{{\color{black}Comparison of classification accuracy of methods based on deep neural networks on ModelNet10 \& 40.}}
\label{table:cmpartsdeep}
\noindent
\thispagestyle{empty}
\vspace*{-\baselineskip}
\centering
%\caption{Table 1. Classification results on ModelNet10 and ModelNet40}
%\begin{tabular}{|p{0.15cm}|c|p{0.5cm}<{\centering}|p{0.5cm}<{\centering}|p{0.7cm}<{\centering}|p{0.7cm}<{\centering}|p{0.7cm}<{\centering}|p{0.7cm}<{\centering}|}
%\colorbox[rgb]{0.6,0.8,1.0}{%
\begin{tabular}{p{0.2cm}<{\centering}cccp{0.001cm}<{\centering}cc}
\toprule[1pt]
\multicolumn{2}{c}{\multirow{2}{*}{Method}}&\multicolumn{2}{c}{ModelNet10}&&\multicolumn{2}{c}{ModelNet40}\\
\cline{3-4} %draw line from column 3 to column 8
\cline{6-7}
\multicolumn{2}{c}{}&Inst.&Class&&Inst.&Class\\
\midrule[0.8pt]
\multirow{11}{*}{\rotatebox{-90}{shape-based}}&3DShapeNets\cite{wuz2015a}&-&83.5&&-&77.3\\
%\cline{2-6}
&VoxNet\cite{maturanad2015a}&-&92.0&&-&83.0\\
%\cline{2-6}
&SubVolSup\cite{qicr2016a}&-&-&&89.2&86.0\\
%\cline{2-6}
&AniProbing\cite{qicr2016a}&-&-&&89.9&85.6\\
%\cline{2-6}
&VRN\cite{brocka2016a}&93.61&-&&91.33&-\\
%\cline{2-6}
&VRN-Ensemble\cite{brocka2016a}&97.14&-&&95.54&-\\
%\cline{2-6}
&FPNN\cite{liy2016a}&-&-&&88.4&-\\
%\cline{2-6}
&PointNet\cite{qicr2016b}&-&-&&89.2&86.2\\
%\cline{2-6}
&PointNet++\cite{qicr2017a}&-&-&&91.9&-\\
%\cline{2-6}
&O-CNN\cite{wangps2017a}&-&-&&90.6&-\\
%\cline{2-6}
&So-Net\cite{lij2018a}&95.7&95.5&&93.4&90.8\\
\hline
\multirow{10}{*}{\rotatebox[origin=c]{-90}{view-based}}
%\cline{2-6}
&DeepPano\cite{shib2015a}&-&88.7&&-&82.5\\
%\cline{2-6}
&PANORAMA-NN\cite{Sfikask2017a}&91.12&-&&90.70&-\\
%\cline{2-6}
&PANORAMA-ENN\cite{Sfikask2017b}&96.85&-&&95.56&-\\
%\cline{2-6}
&MVCNN\cite{suh2015a}&-&-&&-&90.1\\
%\cline{2-6}
&MVCNN-Alex\cite{qicr2016a}&-&-&&92.0&89.7\\
%\cline{2-6}
&MVCNN-MultiRes\cite{qicr2016a}&-&-&&93.8&91.4\\
%\cline{2-6}
&Pairwise\cite{johnse2016a}&94.0&-&&92.0&-\\
%\cline{2-6}
&DomSetClust\cite{wangc2017a}&-&-&&93.8&92.8\\

&RotationNet\cite{kanezakia2018a}&98.46&-&&97.37&-\\
%\cline{2-6}
&VERAM&{95.5}&{95.3}&&93.7&92.1\\
\hline
%\rotatebox{-90}{mix\quad}&\tabincell{c}{\\FusionNet}&\tabincell{c}{\\20,60}&\tabincell{c}{\\\(\times\)}&\tabincell{c}{\\93.1}&\tabincell{c}{\\-}&\tabincell{c}{\\90.8}&\tabincell{c}{\\-}\\
%\multirow{2}{*}{\rotatebox{-90}{mix}}&\multirow{2}{*}{FusionNet}&\multirow{2}{*}{20,60}&\multirow{2}{*}{\(\times\)}&\multirow{2}{*}{93.1}&\multirow{2}{*}{-}&\multirow{2}{*}{90.8}&\multirow{2}{*}{-}\\
%&&&&&&&\\
mix&FusionNet\cite{hegdev2016a}&93.1&-&&90.8&-\\
\bottomrule[1pt]
\end{tabular}
%\caption{Comparison of Classification results on ModelNet10 and ModelNet40.}
\end{table}

{\color{black}
Among shape-based methods, VRN-Ensemble \cite{brocka2016a} achieves the best instance-level accuracy,  $97.14\%$ on ModelNet10 and $95.54\%$ on ModelNet40. This result is by summing predictions from separately trained five VRN models and one Voxception model.
%VRN-Ensemble is the only one among model-based methods whose performance obviously surpasses VERAM.
 When comparing VERAM with the single VRN model, VERAM gets the better performance, $95.5\%$ vs. $93.61\%$ on ModelNet10, and $93.7\%$ vs. $91.33\%$ on ModelNet40.

 Among view-based methods, DeepPano \cite{shib2015a}, PANORAMA-NN\cite{Sfikask2017a} and PANORAMA-ENN\cite{Sfikask2017b} are based on the panoramic image of 3D shape. PANORAMA-ENN has a clear advantage over VERAM, $96.85\%$ vs. $95.5\%$ on ModelNet10, $95.56\%$ vs. $93.7\%$ on ModelNet 40. PANORAMA-ENN needs to obtain three panoramas from different principle axes and each panorama needs to extract SDM, NDM and magnitude of gradient image of NDM, while VERAM only takes grayscale image as input. Moreover, these three panoramic methods can be regarded as based on continuous views, while VERAM and other view-based methods are based on discrete views.

Table~\ref{table:cmpartsview} gives the detail comparison against state-of-the-art discrete view-based methods with their different processing strategies. Apparently, RotationNet, DomSetClust and MVCNN-MultiRes get the better performance.
%Table ~\ref{table:cmpartsview} gives the detail comparison against state-of-the-art discrete view-based methods with their different processing strategies. On ModelNet10, VERAM with AlexNet achieves $95.5\%$ instance-level accuracy and $95.3\%$ class-level accuracy. On ModelNet40, VERAM with AlexNet achieves $93.7\%$ instance-level accuracy and $92.1\%$ class-level accuracy.  Apparently, RotationNet, DomSetClust and MVCNN-MultiRes get the better performance.
By augmenting the classification task with pose estimation, RotationNet gets instance-level accuracy $98.46\%$ on ModelNet10 and $97.37\%$ on ModelNet40, both are state-of-the-art performance. They are achieved by alerting 11 different camera system orientations. According to table 7 of \cite{kanezakia2018a}, with AlexNet, On ModelNet 40, there are 8 of total 11 camera system orientations (except $2$th, $3$th, $4$th) under which the performance of RotionalNet is less than $93.04\%$, which is inferior to VERAM ($93.7\%$). Considering VERAM only uses a single camera system orientation, there is potential for VERAM to further reduce the gap.

%By augmenting the classification task with pose estimation, RotationNet gets instance-level accuracy $98.46\%$ on ModelNet10 and $97.37\%$ on ModelNet40, both are state-of-the-art performance. They are achieved by alerting 11 different camera system orientations. According to table 7 of \cite{kanezakia2018a}, with AlexNet, on ModelNet10, there are 11 of total 12 camera system orientations (except $2$th) under which the performance of RotationNet is less than $94.72\%$, which is inferior to VERAM ($95.5\%$). On ModelNet 40, there are 8 of total 12 camera system orientations (except $2$th, $3$th, $4$th) under which the performance of RotionalNet is less than $93.04\%$, which is also inferior to VERAM ($93.7\%$). Considering VERAM only uses a single camera system orientation, there is potential for VERAM to reduce the gap.

DomSetClust ever gets instance-level accuracy $93.8\%$ and class-level accuracy $92.8\%$ on ModelNet40, and both better than the performance of VERAM. However, to obtain such performance, DomSetClust needs to take grayscale, depth and surface normals image as input and fine-tune the CNN network. When DomSetClust and VERAM both use grayscale image as input, VERAM has an clear advantage, $93.7\%$ vs. $92.2\%$ instance-level accuracy, and $92.1\%$ vs. $91.5\%$ class-level accuracy.

%The comparison of VERAM and MVCNN-Alex is more fair for they both taking grayscale image as input, AlexNet as CNN, and with single resolution. VERAM has an obvious advantage over MVCNN-Alex, $93.7\%$ vs. $92.0\%$ instance-level accuracy and $92.4\%$ vs. $89.7\%$ class-level accuracy. The instance-level accuracies of VERAM and MVCNN-MultiRes  are $93.7\%$ and $93.8\%$, while the class-level accuracies are $92.1\%$ and $91.4\%$ respectively. Their performances are very close but MVCNN-MultiRes uses three resolutions.

The comparison of VERAM and MVCNN-Alex is more fair for they both taking grayscale image as input, AlexNet as CNN, and with single resolution. VERAM has an obvious advantage over MVCNN-Alex, $93.7\%$ vs. $92.0\%$ instance-level accuracy and $92.1\%$ vs. $89.7\%$ class-level accuracy.
The performances of VERAM is matchable with MVCNN-MultiRes, but the latter needs to implement two times of voxelization and three times of rendering.
%{\color{red}\sout{MVCNN-MultiRes needs to implement two times of voxelization and three times of rendering.}}

%The instance-level accuracies of MVCNN-MultiRes and VERAM are $93.8\%$ and $93.7\%$ respectively, MVCNN-Res is better. The class-level accuracies of MVCNN-MultiRes and VERAM are $91.4\%$ and $92.1\%$, and VERAM achieves the better results. Do not forget that VERAM only uses a single resolution of 3D shape, while MVCNN-MultiRes uses three resolutions.MVCNN-MultiRes needs to implement two times of voxelization and three times of rendering.The comparison of VERAM and MVCNN-Alex is more fair for they both use grayscale image, AlexNet, and single resolution. In such circumstance, VERAM has an obvious advantage over MVCNN-Alex, $93.7\%$ vs. $92.0\%$ instance-lever accuracy and $92.4\%$ vs. $89.7\%$ class-level accuracy.

Based on the above study, it can be concluded that under the equivalent conditions of render representation, resolution, CNN architecture and number of views, VERAM outperforms all state-of-the-art view-based methods.
%Furthermore, the performance of VERAM with AlexNet is a little better than ResNet on ModelNet40, but the situation is the reverse on ModelNet10. In sum, the performance difference between VERAM and ResNet is marginal.

\color{black}

\renewcommand{\multirowsetup}{\centering}
\begin{table*}
\caption{{\color{black}Detailed comparison against state-of-the-art discrete view-based methods on ModelNet10 \& ModelNet40.}}
\label{table:cmpartsview}
\noindent
\centering
%\caption{Table 1. Classification results on ModelNet10 and ModelNet40}
%\begin{tabular}{|p{0.15cm}|c|p{0.5cm}<{\centering}|p{0.5cm}<{\centering}|p{0.7cm}<{\centering}|p{0.7cm}<{\centering}|p{0.7cm}<{\centering}|p{0.7cm}<{\centering}|}
%\colorbox[rgb]{0.6,0.8,1.0}{%
\begin{tabular}{cccccccccp{0.0001cm}<{\centering}cccc}
\toprule[1pt]
\multirow{2}{*}{Method}&\multirow{2}{*}{Input}&\multirow{2}{*}{\tabincell{c}{Reso-\\lution}}&\multirow{2}{*}{ CNN}&\multirow{2}{*}{\tabincell{c}{Fine\\tune}}&\multicolumn{4}{c}{ModelNet10}&&\multicolumn{4}{c}{ModelNet40}\\
\cline{6-9}
\cline{11-14}
&&&&&View&Inst.&View&Class&&View&Inst.&View&Class\\
\midrule[0.8pt]
MVCNN\cite{suh2015a}&Gray&1&VGG-M&$\surd$&-&-&-&-&&-&-&80&90.1\\
%\cline{2-6}
MVCNN-Alex\cite{qicr2016a}&Gray&1&AlexNet&$\surd$&-&-&-&-&&20&92.0&20&89.7\\
%\cline{2-6}
MVCNN-MulRes\cite{qicr2016a}&Gray&3&AlexNet&$\surd$&-&-&-&-&&20&93.8&20&91.4\\
%\cline{2-6}
Pairwise\cite{johnse2016a}&Gray+Depth&1&VGG-M&$\surd$&12&94.0&-&-&&12&92.0&-&-\\
%\cline{2-6}
DomSetClust\cite{wangc2017a}&Gray&1&VGG-M&\(\times\)&-&-&-&-&&12&91.9&12&90.4\\
DomSetClust\cite{wangc2017a}&Gray+Depth+Surf&1&VGG-M&\(\times\)&-&-&-&-&&12&93.3&12&92.1\\
DomSetClust\cite{wangc2017a}&Gray&1&VGG-M&$\surd$&-&-&-&-&&12&92.2&12&91.5\\
DomSetClust\cite{wangc2017a}&Gray+Depth+Surf&1&VGG-M&$\surd$&-&-&-&-&&12&93.8&12&92.8\\
%\cline{2-6}
RotationNet\cite{kanezakia2018a}&Gray&1&AlexNet&$\surd$&20&98.46&-&-&&20&97.37&-&-\\

%\cline{2-6}
VERAM&Gray&1&AlexNet&\(\times\)&9&95.5&9&95.3&&9&93.7&9&92.1\\
VERAM&Gray&1&ResNet&\(\times\)&9&96.3&9&96.1&&9&93.2&9&91.5\\
\bottomrule[1pt]
\end{tabular}
%\caption{Comparison of Classification results on ModelNet10 and ModelNet40.}
\end{table*}
\vspace{1ex}

\vspace{8pt}
\noindent\textbf{Comparison with alternative view-selection methods.}
VERAM falls into the category of active view selection. Among the methods in table~\ref{table:cmpartsdeep}, both 3DShapeNets \cite{wuz2015a} and Pairwise \cite{johnse2016a} also adopt active view selection for 3D shape classification. Besides, MV-RNN \cite{xuk2016a} extends RNN-based visual attention model by integrating  MVCNN \cite{suh2015a} into the architecture for 3D object retrieval. To give a comprehensive comparison, we implemented MV-RNN for 3D shape classification and the input to MV-RNN is
%{\color{red}\sout{same as VERAM, i.e.,}}
 $2048$ vector for each rendered image extracted from ResNet152.
 %{\color{red}\sout{A separated CNN2 of performing max pooling across all views is needed to train for MV-RNN. CNN2 achieves $95.27\%$ instance-level accuracy and $95.01\%$ class-level accuracy on ModelNet10, $92.79\%$ instance-level accuracy and $90.38\%$ class-level accuracy on ModelNet40, which are fairly good but still inferior to that of VERAM.}}
 Both VERAM and MV-RNN take grayscale image as input, and 3DShapeNets takes depth image as input. In contrast, Pairwise exploits more information and takes both grayscale and depth image as input. %{\color{red}\sout{Sphere is the sampling space of viewpoints for all these four methods. }} Fig.~\ref{fig:cmpnbv} shows the instance-level accuracy of the four methods on ModelNet10 and ModelNet40 with view $3$, $6$ and $12$. The data of 3DShapeNets and Pairwise is provided by \cite{johnse2016a}.
\begin{figure}[h]
\centering
\includegraphics[width = .48\textwidth]{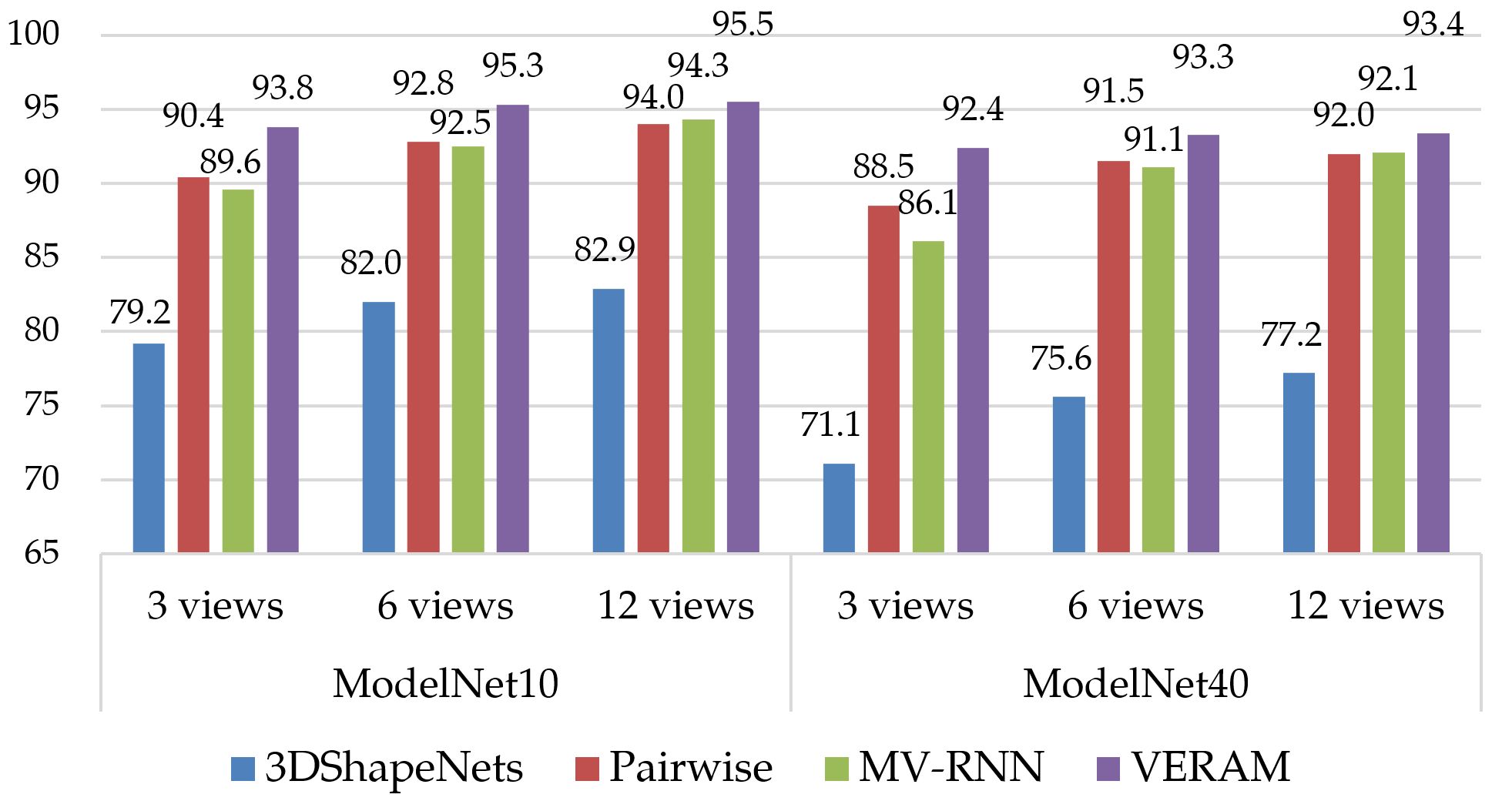}
\caption{Comparison of instance-level accuracy of 3DShapeNets \cite{wuz2015a}, Pairwise \cite{johnse2016a}, MV-RNN \cite{xuk2016a} and VERAM with different number of views on ModelNet10 and ModelNet40.}
\label{fig:cmpnbv}
\end{figure}

3DShapeNets provides the baseline performance on both datasets among all methods. With the deep neural networks pre-trained on ImageNet, Pairwise makes a great progress on the performance. With 12 views, it gets $94.0\%$ and $92.0\%$ instance-level accuracy on ModelNet10 and ModelNet40 respectively. MV-RNN adopts CNN2 of max pooling to perform classification and the accuracy increases with the number of views. The performance of MV-RNN outperforms that of Pairwise with $12$ views. VERAM outperforms 3DShapeNets, Pairwise and MV-RNN on both datasets in every view. Although VERAM also uses the pre-trained model on ImageNet. However, the next view selection of VERAM is embedded seamlessly in RNN with view confidence and view location constrains learning.
\iffalse
{\color{red}\sout{It also can be seen that with $3$ views, the performance of 3DShapeNets, Pairwise and MV-RNN are fairly low and they need to collect more views to further enhance their performance. By contrast, the performance of VERAM can quickly converge with only a limited number of views. It should be noticed that in ModelNet10, the classification accuracy of VERAM with only}
%{\color{red}\sout{$3$}}
{\color{black}$6$} views is better than that of Pairwise and MV-RNN with $12$ views (
%{\color{red}\sout{$95.8$}}
{\color{black}$95.6\%$} vs. $94.0\%$ and
%{\color{red}\sout{$95.8\%$}}
{\color{black}$95.6\%$} vs. $94.3\%$).
\fi

{\color{black}\subsection{Evaluation of the enhancement of VERAM}
%\vspace{8pt}

\noindent In this subsection, we will evaluate the three key technical components of VERAM on conducting to improve the performance step by step. ResNet is used to encode the rendered image for it is more compact. We use linear mapping for recurrent network instead of LSTM to stress the ability of three key components to select more discriminative views to enhance the view learning. In the end of this subsection, we will present the results when using LSTM as recurrent network. We trained $5$ models for each network setting with the specified super parameters, and use the average class-level accuracy for comparison.}
\vspace{8pt}

\noindent\textbf{Enhancing the information flow of gradient.}
RNN-based visual attention model suffers from the problem of  unbalanced training of the subnetworks, and the estimated view locations tend to get stuck at the boundary in the view parameterization space. Fig.~\ref{fig:locdistribute} (left) shows the heat map of view location frequency of each time step by applying  the visual attention model presented in \cite{mnihv2014a}, denoted as ClassicalRAM, to predict all $chairs$ in the testing set of ModelNet10, time steps $T$=$4$. It can be seen that starting from the view located at $(6, 4)$, the views of the next three steps almost all locate at the boundary $(1, 1)$.
%{\color{red}\sout{It is obvious that the view estimation subnetwork of this model is poorly trained.}}
To solve this issue, three schemes are proposed for VERAM as described in subsection 3.2.2 to enhance information flow of gradient backpropagation. For simplicity, we call it BoundaryRAM. Fig.~\ref{fig:locdistribute} (right) shows the heat map of view location frequency of each time step by applying a trained BoundaryRAM model to the same $chair$ dataset. There are significant difference among the heat maps of different time steps.
\begin{figure}[h]
\centering
{\includegraphics[width = .22\textwidth]{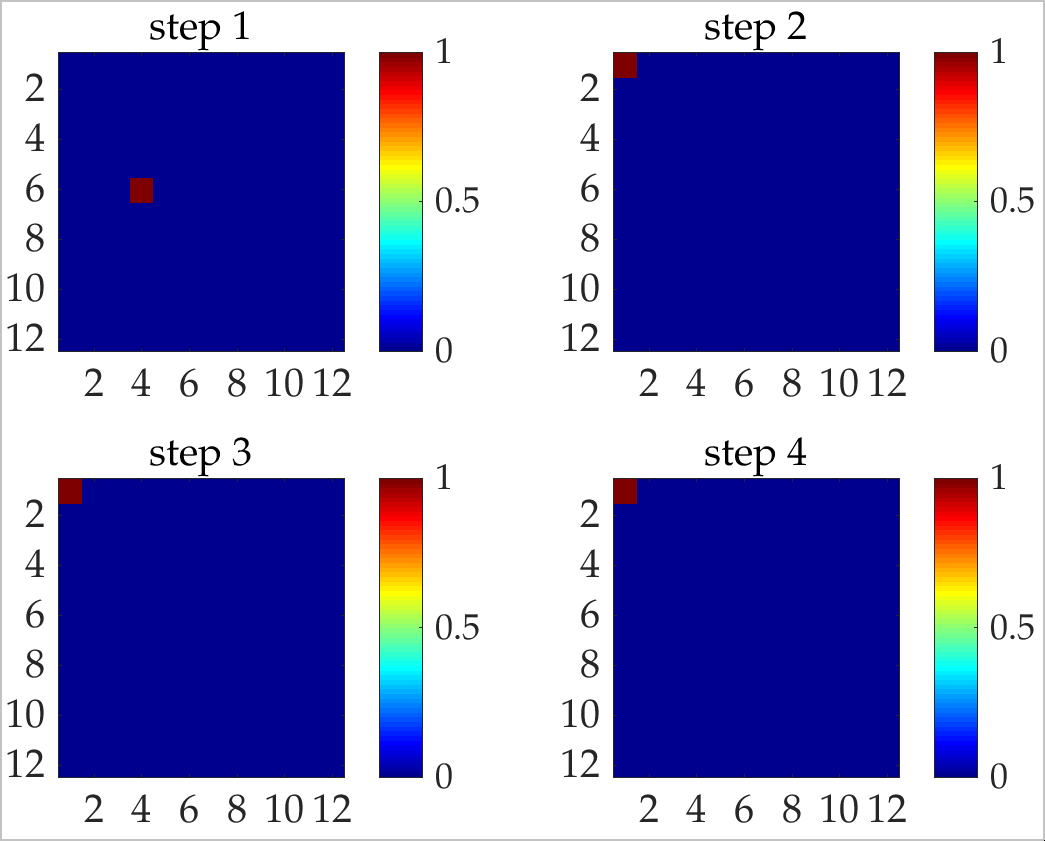}\quad\includegraphics[width = .22\textwidth]{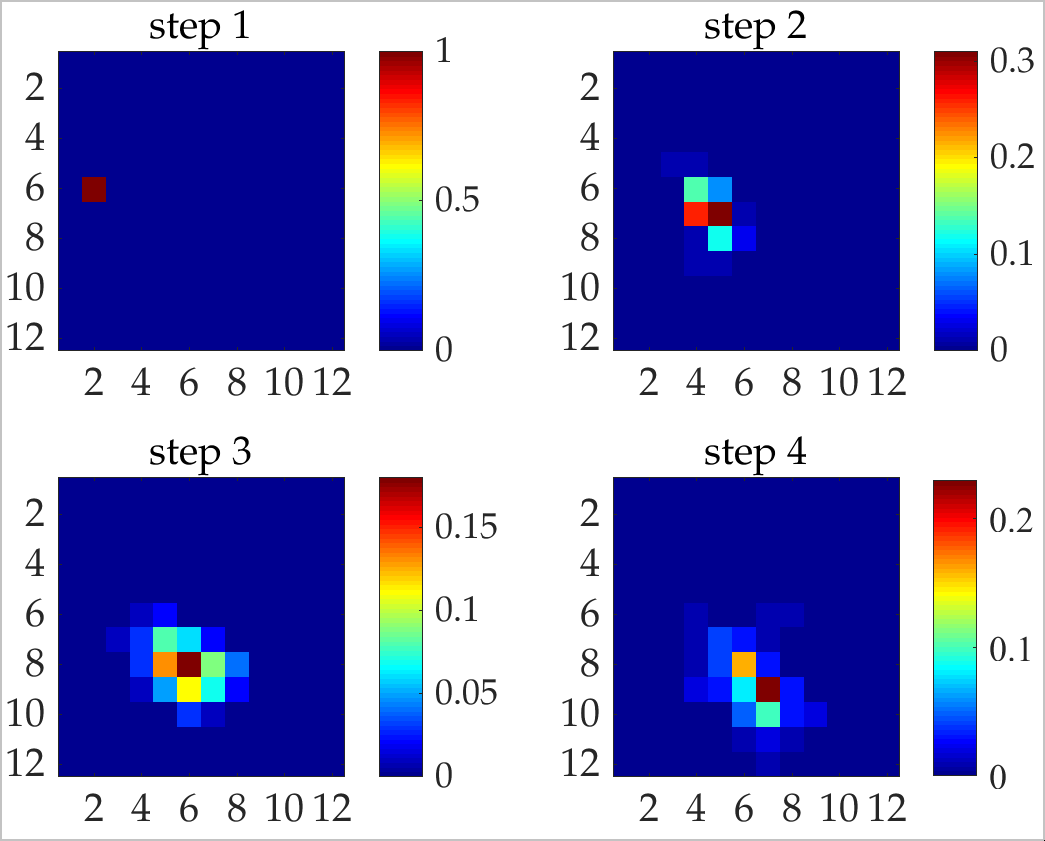}}

\caption{Heat maps of view location frequency of predicating all $chairs$ in the testing set of ModelNet10, time steps $T$=$4$. Left, by using ClassicalRAM \cite{mnihv2014a}. Right, by using BoundaryRAM. }
\label{fig:locdistribute}
\end{figure}

%{\color{red}\sout{Training RNN-based visual attention model involves stochastical components, including parameter initialization, dropout operation, REINFORCE, etc., so the performance of the same network with the same initial parameters may fluctuate among different trained models. For objective evaluation, we trained $5$ models for each network setting with the specified super parameters, and use the average class-level accuracy for comparison.}}
 %as shown in Fig.~\ref{fig:cmpboundaryram}, Fig.~\ref{fig:cmpconfram} and Fig.~\ref{fig:cmplocram}, .

The comparison of the average class-level accuracy between ClassicalRAM \cite{mnihv2014a} and BoundaryRAM on ModelNet10 and ModelNet40 is shown in Fig.~\ref{fig:cmpboundaryram}. The horizontal axis is time steps $T$ from $1$ to $6$. Note that $T$ is a super parameter, and the accuracy of each $T$ is the average value from $5$ trained models with the same network setting. Leaving out $T$=$1$, i.e., $T$=$2\sim6$, the accuracy of BoundaryRAM goes up from min $1.80\%$ to max $5.37\%$ on ModelNet10, and from min $0.65\%$ to max $2.76\%$ on ModelNet40. From Fig.~\ref{fig:locdistribute} and Fig.~\ref{fig:cmpboundaryram}, it can be seen that the three critical schemes presented in subsection 3.2.2 can effectively overcome the issue that the estimated views getting stuck at the boundary, and achieve a apparent performance enhancement.
\begin{figure}[h]
\centering

{\includegraphics[width = .23\textwidth]{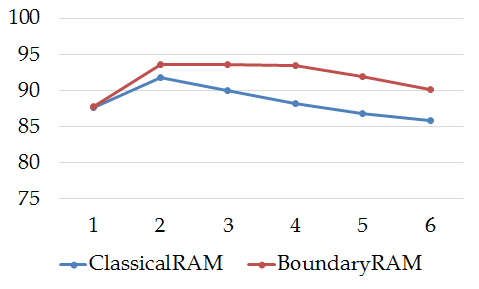}\quad\includegraphics[width = .23\textwidth]{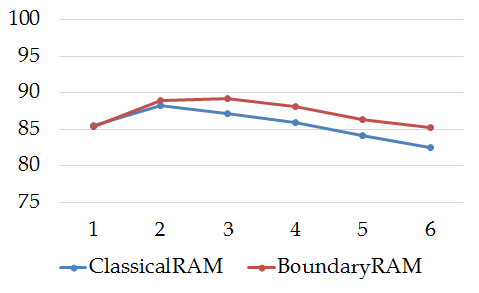}}
\caption{Comparison of average class-level accuracy between ClassicalRAM \cite{mnihv2014a} and BoundaryRAM. Left, on ModelNet10. Right, on ModelNet40. }
\label{fig:cmpboundaryram}
\end{figure}

\vspace{8pt}
\noindent\textbf{Learning with view confidence.}
The classification subnetwork of VERAM is easier to train than the view estimation one, and learned attention policy can be easily trapped in local optimization. In practice, with the same network, we find the accuracy of each category from different trained BoundaryRAM model fluctuates widely.
%{\color{red}\sout{In table~\ref{table:unstalble}, columns from $2$ to $3$ show the min, max, mean and variance of the class-level accuracy of the $5$ trained BoundaryRAM models with $T$=$4$ on ModelNet10, and columns from $4$ to $5$ show that of the other $5$ trained BoundaryRAM models with $T$=$6$. When the time step $T$ is set to $4$, the average class-level accuracy is $93.6\%$ and the average variance is $1.66$. However, when the number of time step $T$ increases to $6$, the average variance rises to $3.26$, and average class-level accuracy drops to $90.1\%$. Obviously, }}
 With the number of time steps increases, the model trained by BoundaryRAM will be more unstable. To alleviate this problem, based on BoundaryRAM, in subsection 3.2.3, we propose a method of learning with view confidence for REINFORCE to provide effective guidance to agent on where to deploy the model's attention, here called ConfRAM.
%{\color{red}In table~\ref{table:unstalble}, the right two columns show the min, max, mean and variable of the average class-level accuracy of $5$ trained ConfRAM models on ModelNet10 with $6$ time steps. ConfRAM achieves $95.0\%$ average class-level accuracy and the variance is only $0.84$. The variance is smaller than that of BoundaryRAM with $4$ time steps, and much smaller than that of BoundaryRAM with $6$ time steps. This indicates that learning with view confidence can force our model to select more discriminative views and avoid trapping in local minima with low discriminative views.}}
  The comparison of the average class-level accuracy among ClassicalRAM \cite{mnihv2014a}, BoundaryRAM and ConfRAM on ModelNet10 and ModelNet40 is shown in Fig.~\ref{fig:cmpconfram}.

\begin{figure}[h]
\centering
{\includegraphics[width = .23\textwidth]{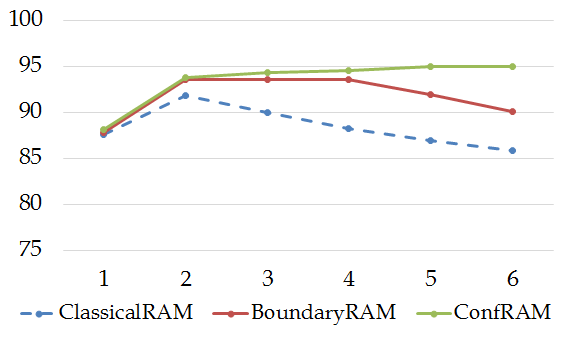}\quad\includegraphics[width = .23\textwidth]{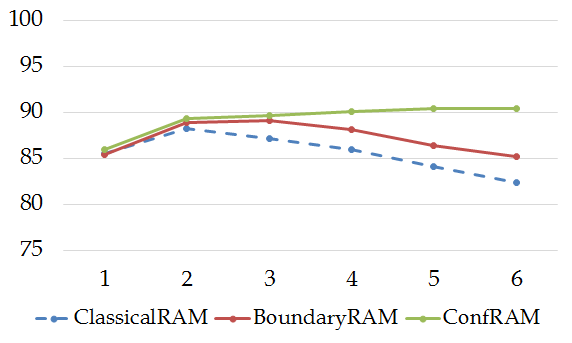}}
\caption{Comparison of average class-level accuracy among ClassicalRAM \cite{mnihv2014a}, BoundaryRAM, and ConfRAM. Left, on ModelNet10. Right, on ModelNet40.  }
\label{fig:cmpconfram}
\end{figure}

From Fig.~\ref{fig:cmpconfram}, it can be seen that, the performance of ClassicalRAM drops after the number of time steps $T$=$2$, while BoundaryRAM drops after $T$=$4$ on ModelNet10 and after $T$=$3$ on ModelNet40. By contrast, ConfRAM can exploit more views and converge at about $T$=$6$ with significant higher performance. Compared with BoundaryRAM, leaving out $T$=$1$, i.e., $T$=$2\sim6$, the accuracy of ConfRAM goes up from min $0.20\%$ to max $4.87\%$ on ModelNet10, and from min $0.36\%$ to max $5.24\%$ on ModelNet40. %Fig.~\ref{fig:confinst} gives a detail example. The 3D shape is $nightstand\_0224$. The first row shows the visited view of each time step when using a trained BoundaryRAM model with $T$=$6$. The predication probability of this shape belonging to $nightstand$ is only $0.39$ and it is misclassified as $dresser$. The second row is the visited view of each time step when using another trained BoundaryRAM model but $T$=$4$. The shape is predicated correctly with probability $0.75$. The third row shows the visited view of each time step when using a trained ConfRAM model. The shape is classified correctly and the probability reaches to $0.93$. This means ConfRAM has collected more positive information for the predication.

%\begin{figure}[h]
%\centering
%{\includegraphics[width = .45\textwidth]{confinst_minor.png}}
%\caption{Visited view of each time step when predicating 3D shape $nightstand\_0224$ in the testing set of ModelNet10, with (a) BoundaryRAM ($T$=$6$), (b) BoundaryRAM ($T$=$4$), and (c) ConfRAM ($T$=$6$).}
%\label{fig:confinst}
%\end{figure}

\vspace{8pt}
\noindent\textbf{Learning with view location constrains.}
ConfRAM can achieve a stable and fairly good performance. However, the visited view location of each time step may overlap. %As shown in Fig.~\ref{fig:confinst} (c), both location (6, 12) and (5, 12) repeat two times.
In subsection 3.2.4, based on BoundaryRAM and ConfRAM, a weak regular term is adopted by VERAM to keep the visited views separated from each other, here called LocRAM, equally to the whole VERAM. The comparison of the average class-level accuracy among ClassicalRAM \cite{mnihv2014a}, BoundaryRAM, ConfRAM and LocRAM on ModelNet10 and ModelNet40 is shown in Fig.~\ref{fig:cmplocram}.
\begin{figure}[h]
\centering
{\includegraphics[width = .23\textwidth]{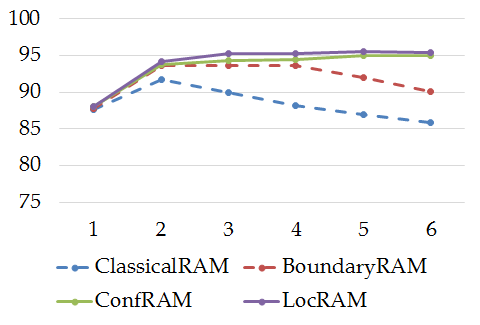}\quad\includegraphics[width = .23\textwidth]{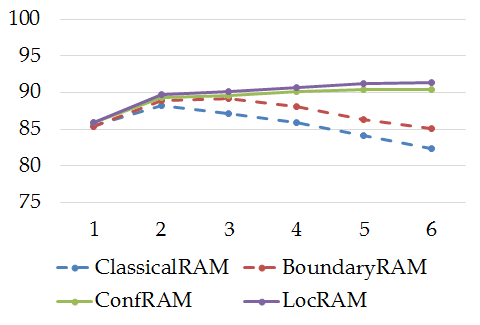}}
\caption{Comparison of average class-level accuracy among ClassicalRAM \cite{mnihv2014a}, BoundaryRAM, ConfRAM and LocRAM. Left, on ModelNet10. Right, on ModelNet40.}
\label{fig:cmplocram}
\end{figure}

Fig.~\ref{fig:cmplocram} shows LocRAM learning with the weak regular term of view location constrains can obtain a clear performance improvement. Compared with ConfRAM, leaving out $T$=$1$, i.e., $T$=$2\sim6$, the accuracy of LocRAM goes up from min $0.40\%$ to max $0.94\%$ on ModelNet10, and from min $0.43\%$ to max $0.91\%$ on ModelNet40. It should be pointed out that the regular term of VERAM is only a weak constrain. %Correspondingly to Fig.11, Fig.~\ref{fig:locinst} shows the visited view of each time step when using a trained LocRAM model with $T$=$6$ for $nightstand\_0224$ in ModelNet10, it can be seen that the regular term has taken effect.

%\begin{figure}[h]
%\centering
%{\includegraphics[width = .45\textwidth]{locinst_minor.png}}
%\caption{Visited view of each time step when predicating 3D shape $nightstand\_0224$ in the testing set of ModelNet10, with LocRAM ($T$=$6$).}
%\label{fig:locinst}
%\end{figure}
{\color{black}The above experiments of this subsection use linear mapping as recurrent subnetwork, and the previous step will have less impact when the interval between which and the last step $T$ increases. When replacing linear mapping with LSTM, the comparison of the average class-level accuracy among ClassicalRAM \cite{mnihv2014a}, BoundaryRAM, ConfRAM and LocRAM on ModelNet10 and ModelNet40 is shown in Fig.~\ref{fig:cmplocramlstm}. Since LSTM is effective at capturing long-term temporal dependencies, the already selected discriminative view prevents the performance from decreasing obviously. However, VERAM with three key components still achieves a significant performance boost over ClassicalRAM. When $T$=$6$, the accuracy is improved by $3.23\%$ on ModelNet10 ($95.54\%$ vs. $92.31\%$), and by $3.12\%$ on ModelNet40 ($91.44\%$ vs. $88.32\%$).}
\begin{figure}[h]
\centering
{\includegraphics[width = .23\textwidth]{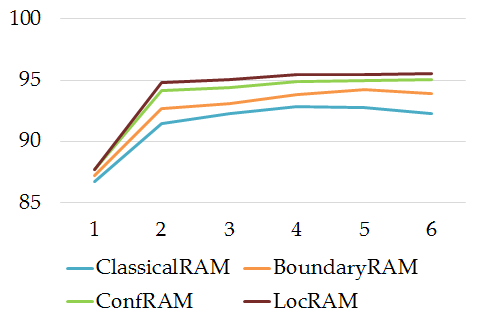}\quad\includegraphics[width = .23\textwidth]{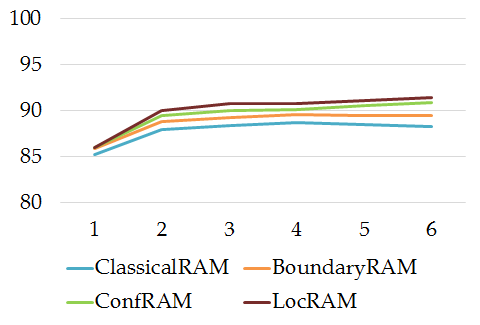}}
\caption{Comparison of average class-level accuracy among ClassicalRAM \cite{mnihv2014a}, BoundaryRAM, ConfRAM and LocRAM when using LSTM as recurrent cells. Left, on ModelNet10. Right, on ModelNet40.}
\label{fig:cmplocramlstm}
\end{figure}

\color{black}{\subsection{Affecting factors on the performance of VERAM}}\color{black}
%\subsection{Comprehensive assessment of VERAM}
%The architecture of VERAM is complex in many aspects. RNN is used for deep reinforcement learning and the performance is affected by time steps $T$, CNN can be embedded in RNN for image feature extraction, etc. In this section, we will present comprehensive assessment of VERAM.
\vspace{8pt}

{\color{black}\noindent\textbf{Effect of different recurrent subnetworks.}
The final predication is based on the hidden state of recurrent subnetwork which is updated over each time step. Linear mapping and LSTM are mostly used for this purpose. When using linear mapping as recurrent subnetwork, formula $(3)$ is specified as $h_t$= $ReLU (Linear( h_{t-1})+o_t))$. For LSTM, the total units is set to $1024$ and each of which is composed of cell, input gate, output gate and forget gate. Fig.~\ref{fig:lstmlinearsupplement} gives the comparison of the performance of VERAM with linear mapping and with LSTM on ModelNet40 when $T$=$3, 6, 9$. The left uses AlexNet as CNN while the right uses ResNet as CNN.

From Fig.~\ref{fig:lstmlinearsupplement}, it can be seen that LSTM achieves the better performance than linear mapping. When using AlexNet to encode the visual representation, VERAM with LSTM has significant advantage over with linear mapping. For example, the instance-level accuracy increases $3.16\%$ from $90.56\%$ to $93.72\%$ by LSTM when $T$=$9$. When using ResNet as CNN, LSTM still achieves the better performance than linear mapping. However, the improvement is slight. When $T$=$9$, it only gets instance-level accuracy $0.52\%$ increment from $92.67\%$ to $93.19\%$. Besides, for both AlexNet and ResNet, the performance of linear mapping when $T$=$9$ is slightly lower than when $T$=$6$, but the performance of LSTM is quite the opposite.

\begin{figure}[h]
\centering
{\includegraphics[width = .48\textwidth]{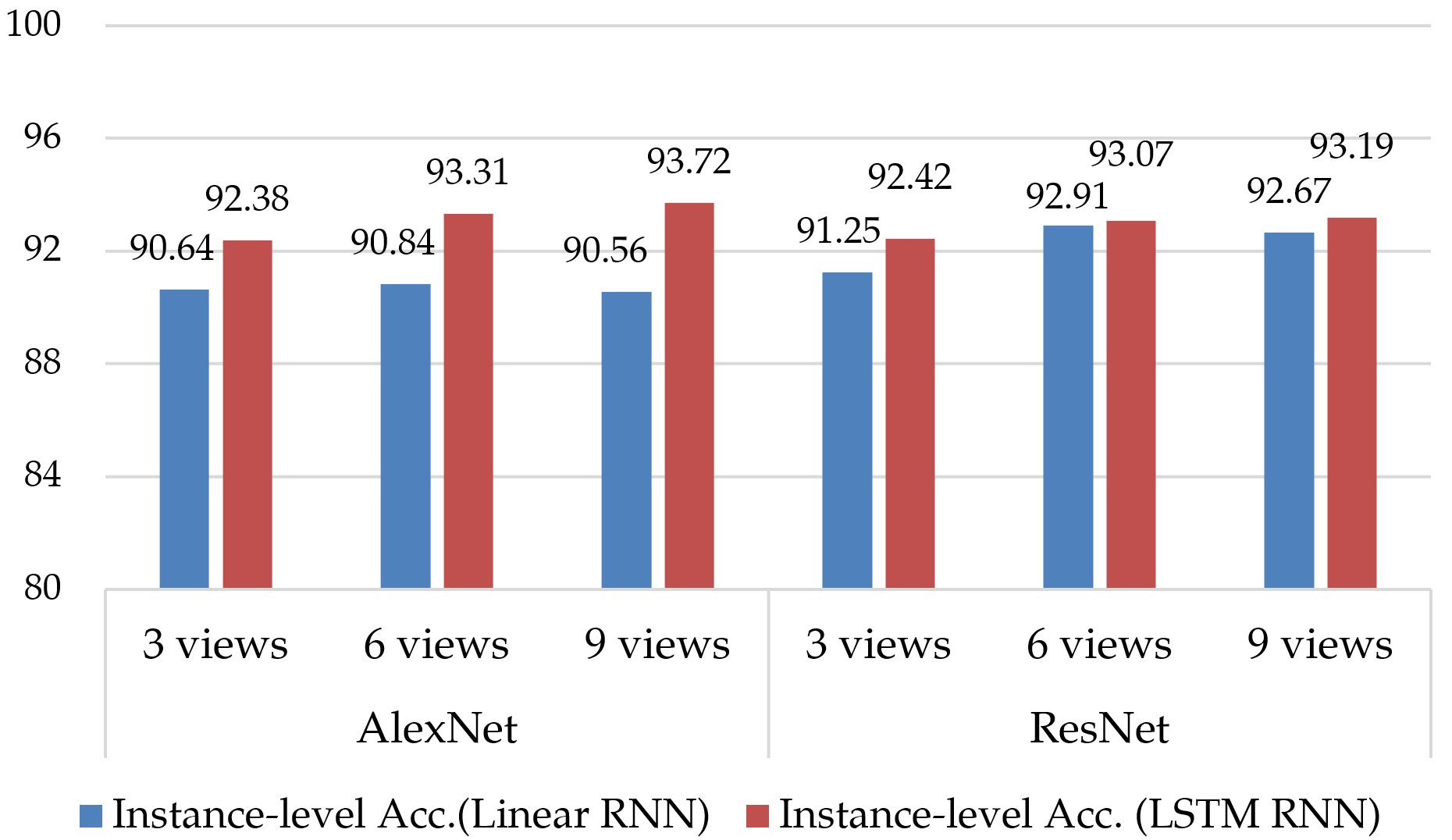}}
\caption{{\color{black}Comparison of the best instance-level accuracy of VERAM with linear mapping and with LSTM on ModelNet40 when $T$=$3, 6, 9$. Left, using AlexNet as CNN. Right, using ResNet as CNN.}}
\label{fig:lstmlinearsupplement}
\end{figure}
}

{\color{black}\noindent\textbf{Effect of different CNN architectures.} As described in subsection 3.1.2, VERAM needs to encode the visual representation for the rendered image at each time step. A number of advanced CNN architectures can be adopted for implementation. Among them, AlexNet is the first notably successful CNN architecture on ImageNet while ResNet is the recently proposed representing the advanced level, and they are compared in this subsection. For each 2D image $x_t$ of 3D shape, we extract features $v_t\in R^{4096}$ from layer $fc6$ of the AlexNet pre-trained on ImageNet. Analogously, we extract features $v_t\in R^{2048}$ from layer $flatten0\_output$ of the ResNet152 pre-trained on ImageNet 11K, which is available on MXNet \mbox{\cite{chent2015a}}. For efficient training, we didn't fine-tune AlexNet or ResNet model on ModelNet. Fig.~\ref{fig:alexvsresnet}
%\footnote{The data of this figure is same with Fig.~\ref{fig:lstmlinearsupplement} for CNN and RNN are tightly coupled in VERAM. }
presents the performance of VERAM with AlexNet and with ResNet on ModelNet40 when $T$=$3, 6, 9$. The left uses linear mapping as recurrent network while
the right uses LSTM. }

\begin{figure}[h]
\centering
{\includegraphics[width = .48\textwidth]{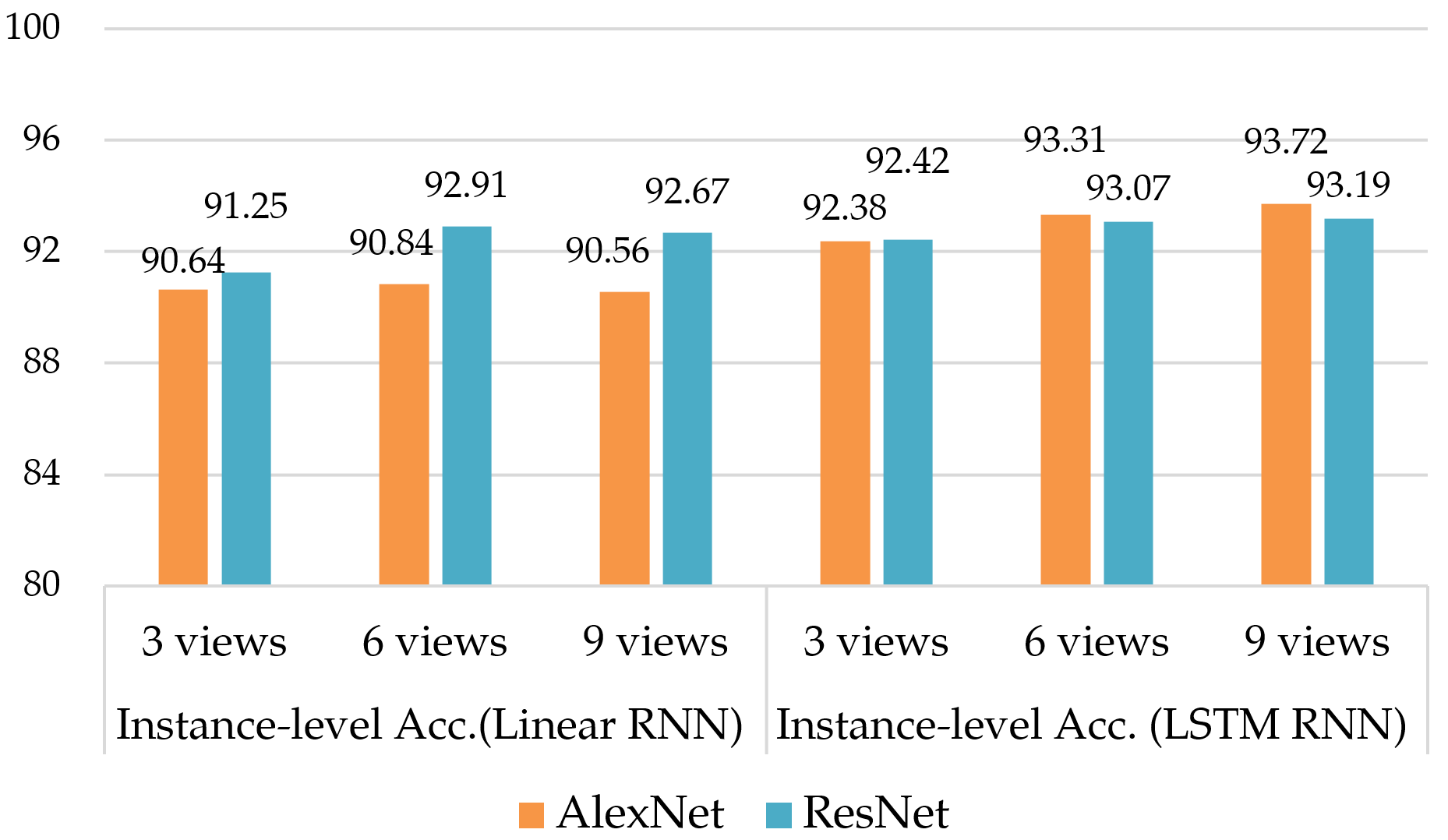}}
\caption{{\color{black}Comparison of the best instance-level accuracy of VERAM with AlexNet and ResNet on ModelNet40 when $T$=$3, 6, 9$. Left, using linear mapping as RNN. Right, using LSTM as RNN.}}
\label{fig:alexvsresnet}
\end{figure}

 {\color{black}Fig.~\ref{fig:alexvsresnet} indicates that when using linear mapping as RNN, the performance of ResNet soars about $2$ percent compared with that of AlexNet and has a clear advantage. However, when using LSTM as RNN, the performance of AlexNet and ResNet are comparable to each other. At $T$=$9$, the best instance-level accuracy of VERAM with AlexNet is $93.72\%$, and with ResNet, it is $93.19\%$ . Theoretically, the feature extracted from ResNet has more capacity of discrimination than feature extracted from AlexNet. However, the features extracted from CNN in each time step are further merged by LSTM, and acoording to Fig.~\ref{fig:lstmlinearsupplement}, the improvement of LSTM on VERAM with AlexNet is much more significant than on VERAM with ResNet. As a result, the performance gap between these two CNNs is reduced. We noticed the performance difference among different CNN architectures in recent work \cite{kanezakia2018a} on 3D shape recognitions is about $1\%$.}}

\begin{figure}[h]
\centering
{\includegraphics[width = .45\textwidth]{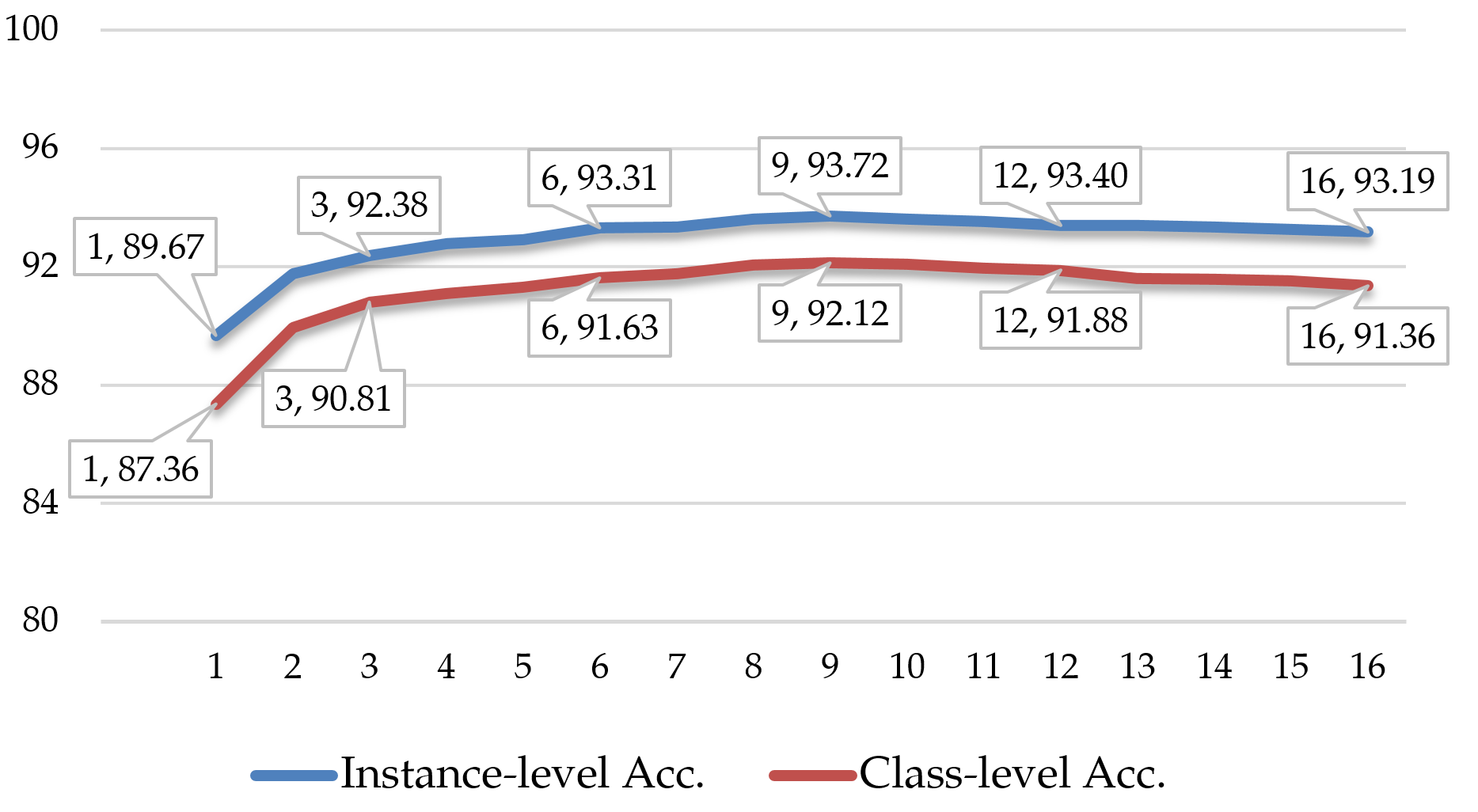}}
\caption{ Best instance-level and class-lever accuracy of VERAM on ModelNet40 with time steps $T$ from 1 to 16. AlexNet is used to encode the visual representation and LSTM is adopt as recurrent network.}
\label{fig:numstepsrevise}
\end{figure}

{\color{black}\noindent\textbf{Effect of time steps T.}  Time steps $T$ in VERAM is a super parameter and it determines how many images does VERAM need to render and observe before emit the classification predication.} {\color{black}Fig.~\ref{fig:numstepsrevise} shows the best instance-level and class-lever accuracy VERAM achieved on ModelNet40 with different time steps $T$ from $1$ to $16$. AlexNet is used to encode the visual representation and LSTM is adopt as recurrent subnetwork. For clarity, we only append $T$ and its accuracy for $1,3,6,9,12,16$. In Fig.~\ref{fig:numstepsrevise}, the rate of accuracy growth is fast in the first few steps, then becomes slower and slower with the increase of time step $T$ and converges at $T$=$9$, where VERAM achieves the best performance, $93.7\%$ instance-level accuracy and $92.1\%$ class-level accuracy. After that, the performance even slightly decreases. From Fig.~\ref{fig:numstepsrevise} it can be concluded as follows:

  \indent 1) {\color{black}The performance of VERAM can quickly converge within a few time steps.} On ModelNet40, VERAM can get $92.38\%$ instance-level accuracy with only 3 views, which is as high as $98.6\%$ of the best performance ($93.72\%$).

  \indent 2) Increasing $T$ would do little to improve the performance after the convergence.} VERAM uses the same parameters and network for each time step, so that the total number of parameters will not expand with the increase of $T$. Although increasing $T$ means VERAM can obtain more information, but the larger $T$ means VERAM needing to cope with more views with the same number of parameters.

%{\color{red}\sout{Fig.~\ref{fig:numstepsrevise} shows the best class-lever accuracy VERAM obtained on ModelNet10 and ModelNet40 with different time steps $T$ from $1$ to $12$. It can be seen that (1) the performance of VERAM can quickly converge within a few time steps, (2) the performance of VERAM is very stable with different $T$ after $T>3$. In our experiments VERAM achieves the best accuracy $96.1\%$ on ModelNet10 with $T$=$7$, and best accuracy $91.5\%$ on ModelNet40 with $T$=$6$.}}

\begin{figure}[h]
\centering
\includegraphics[width = .45\textwidth]{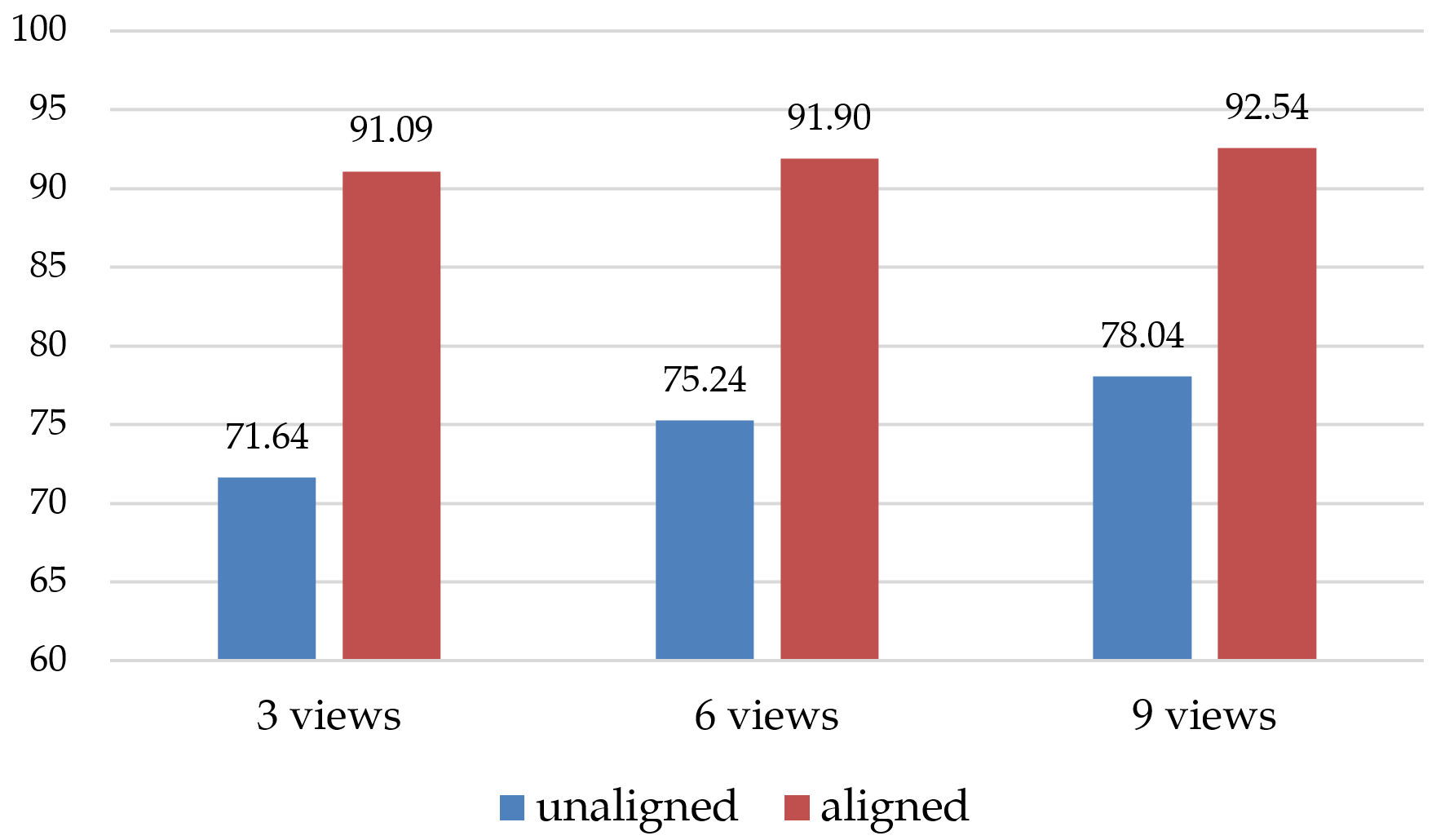}
\caption{Comparison of the instance-level accuracy of VERAM on unaligned and aligned test dataset of ModelNet40.}
\label{fig:alignment}
\end{figure}

\vspace{8pt}
\noindent\textbf{Effect of shape alignment.} {\color{black}
%VERAM is an view-based method and in each time step, it actively chooses the next viewpoint according to the current hidden units.
As Pairwise \cite{johnse2016a} and RotationNet \cite{kanezakia2018a}, VERAM needs a unified coordinate system to render shapes and each shape should be rendered from pre-defined viewpoints, so shape alignment is important to the proposed method. On the website of ModelNet \cite{wuz2015a}, all shapes in ModelNet10 are manually cleaned and their orientations are well aligned. The shapes in ModelNet40 are also cleaned, but a small part of shapes are needed to align in the pre-processing stage. The reported performance of VERAM is based on such aligned orientation.

To quantify the effect of the alignment on the performance, we conduced an experiment with aligned ModelNet40 newly released on the website of ModelNet. Taking AlexNet as CNN and LSTM as RNN, we trained three VERAM models ($T$=$3$, $T$=$6$ and $T$=$9$) with the aligned training dataset and tested the classification on both aligned and unaligned (i.e., randomly rotated) shapes. The results are shown in Fig.~\ref{fig:alignment}. The instance-level accuracy of unaligned shapes only got $71.64\%$, $75.24\%$ and $78.04\%$ with $3$, $6$ and $9$ views.
%which are much lower than performance achieved on the aligned shapes.
%,i.e. $91.09\%$, $91.90\%$ and $92.54\%$
Moreover, it only increases slightly with the increment number of views. These indicate that VERAM is rather sensitive to the pre-defined viewpoints. We notice that this limitation is generally encountered in several view-based methods. For instance, RotationNet\cite{kanezakia2018a} achieves $38\%$ on ShapeNetCore55 dataset. In future, we plan to merge alignment network into VERAM to alleviate this problem.}

%{\color{red}\sout{VERAM falls into the category of active view selection. In each time step, the agent actively chooses the next viewpoint according to the current hidden units. Shape alignment is important to the proposed method. On the ModelNet website, all shapes in ModelNet10 are manually cleaned and their orientations are well aligned. The shapes in ModelNet40 are cleaned but not rigidly aligned, but still, almost all shapes in this subset satisfy the upright assumption. The reported performance of VERAM is based on such canonical orientation.}}

%{\color{red}\sout{We also aligned all the shapes in ModelNet40 rigidly, but have not obtained perceptible performance improvement. We suspect a few not rigidly aligned data may benefit the learning to prevent overfitting.}}

\vspace{8pt}
\noindent\textbf{Complexity.}
{\color{black}With the extracted features from AlexNet and using LSTM as RNN}, the total number of parameters of VERAM is about $12.8$M,
%{\color{red}\sout{8M}}
which is lesser than that of VGG16 ($14.7$M) and ResNet152 ($60$M). In our single TITAN X GPU platform and in the testing phase, it takes less than $0.003$ second to extract features for an image by AlexNet. Besides feature extraction, rendering and predication run very fast, and it only needs about $0.005$ second to emit the decision. As shown in Fig.~\ref{fig:numstepsrevise}, the performance of VERAM is very stable with different $T$ after $T > 3$. In summary, if we set $T$=$6$, which can already obtain high accuracy, it takes about $0.025$ second to classify a 3D shape (not including the time spent on moving the camera).

\iffalse
{\color{red}
\vspace{8pt}
***this part is removed to supplement material for page limitation.***\sout{
\noindent\textbf{Effect of enhancing baseline voting.}
With the same image representation extracted from ResNet152, we compared the performance of VERAM with that of the baseline image voting. As shown in Fig.~\ref{fig:vote}, on ModelNet10, VERAM has results in an enhancement by $4.1\%$ ($96.1\%$ vs. $92.0\%$). On ModelNet40, VERAM has results in an enhancement by $6.2\%$ ($91.5\%$ vs. $85.3\%$). Furthermore, to get the best performance, VERAM only needs $7$ views on ModelNet10 and $6$ views on ModelNet40, while the baseline voting needs all 144 views. VERAM has got an obvious enhancement on both effectiveness and efficiency.}
\begin{figure}[h]
\centering
\includegraphics[width = .35\textwidth]{voting.png}
\caption{{\color{red}\sout{Class-level accuracy of baseline image voting and VERAM on ModelNet10 and ModelNet40.}}}
\label{fig:vote}
\end{figure}}

\fi

{\color{black}{\subsection{Detailed statistics}}
%\vspace{8pt}
%\noindent\textbf{Detailed statistics.}

%{\color{red}\sout{VERAM achieves average class-level accuracy $96.1\%$ on ModelNet10 and $91.5\%$ on Modelnet40. The detail class-level accuracy of each category is reported in table~\ref{table:detailclassresult}}}.

{\color{black}\noindent VERAM achieves the best average class-level accuracy $96.1\%$ on ModelNet10 (with ResNet) and {\color{black}$92.1\%$} on Modelnet40 (with AlexNet). The number of correctly classified shapes over the total number of shapes of each category with these two best models are reported in table~\ref{table:bestclasslevelresult}.
}

  %{\color{red}\sout{On ModelNet10, $nightstand$ and $dresser$ are the two worst categories and they both achieve $89.5\%$ class-level accuracy. The number of shapes in the testing set of $nightstand$ is $86$. Total $9$ shapes of $nightstand$ are misclassified, and among them, $2$ shapes are misclassified as $table$ and $7$ shapes are misclassified as $dresser$. There are also $9$ shapes of $dresser$ that have been misclassified, one of them is misclassified as $desk$ and the others are misclassified as $nightstand$. Two misclassified shapes of $nightstand$ and $dresser$ with their category probability are shown in the first row of Fig.~\ref{fig:misclass}.}}

{\color{black}
   On ModelNet10, $nightstand$ and $table$ are the two worst categories and they achieve $86.05\%$  and $92\%$ class-level accuracy respectively. $12$ out of $86$ $nightstands$ in the testing set are misclassified. One each of them is misclassified as $desk$ and $table$, and the other $10$ $nightstands$ are misclassified as $dresser$. A total of $8$ $tables$ are misclassified. One each of them is misclassified as $bathhub$ and $dresser$ and the others are misclassified as $desk$. Two misclassified shapes of $nightstand$ and $table$ are shown in the first row of Fig.~\ref{fig:misclassrevise}.}

\begin{table}[h]\footnotesize
\caption{{\color{black}The number of correctly classified shapes over the total number of shapes of each category on ModelNet10 and ModelNet40 with the best class-level accuracy VERAM.}}
\label{table:bestclasslevelresult}
\noindent
%\caption{Table 1. Classification results on ModelNet10 and ModelNet40}
%\begin{tabular}{|p{0.15cm}|c|p{0.5cm}<{\centering}|p{0.5cm}<{\centering}|p{0.7cm}<{\centering}|p{0.7cm}<{\centering}|p{0.7cm}<{\centering}|p{0.7cm}<{\centering}|}
\thispagestyle{empty}
\vspace*{-\baselineskip}
%\colorbox[rgb]{0.6,0.8,1.0}{
\begin{tabular}[z]{@{}llcccc}
\toprule[1pt]
\multirow{2}{*}{\rotatebox[origin=c]{270}{ModelNet10}}&bathtub&bed&chair&desk&dresser\\
&48/50&99/100&100/100&83/86&81/86\\
\cline{2-6}
&monitor&\textbf{nightstand}&sofa&\textbf{table}&toilet\\
&100/100&\textbf{74/86}&97/100&\textbf{92/100}&100/100\\
\midrule[0.8pt]
\multirow{16}{*}{\rotatebox[origin=c]{-90}{ModelNet40}}&airplane&bathtub&bed&bench&bookshelf\\
&100/100&47/50&100/100&16/20&98/100\\
\cline{2-6}
&bottle&bowl&car&chair&cone\\
&93/100&17/20&100/100&98/100&19/20\\
\cline{2-6}
&\textbf{cup}&curtain&desk&door&dresser\\
&\textbf{12/20}&19/20&81/86&19/20&78/86\\
\cline{2-6}
&flowerpot&glassbox&guitar&keyboard&lamp\\
&17/20&94/100&100/100&20/20&18/20\\
\cline{2-6}
&laptop&mantel&monitor&\textbf{nightstand}&person\\
&20/20&96/100&100/100&\textbf{62/86}&20/20\\
\cline{2-6}
&piano&plant&radio&rangehood&sink\\
&95/100&95/100&19/20&93/100&18/20\\
\cline{2-6}
&Sofa&stairs&stool&table&tent\\
&97/100&18/20&18/20&79/100&19/20\\
\cline{2-6}
&toilet&tvstand&vase&wardrobe&xbox\\
&100/100&92/100&89/100&18/20&16/20\\

\bottomrule[1pt]
\end{tabular}
\end{table}
\vspace{4pt}

\begin{figure}[h]
\centering
{\includegraphics[width = .45\textwidth]{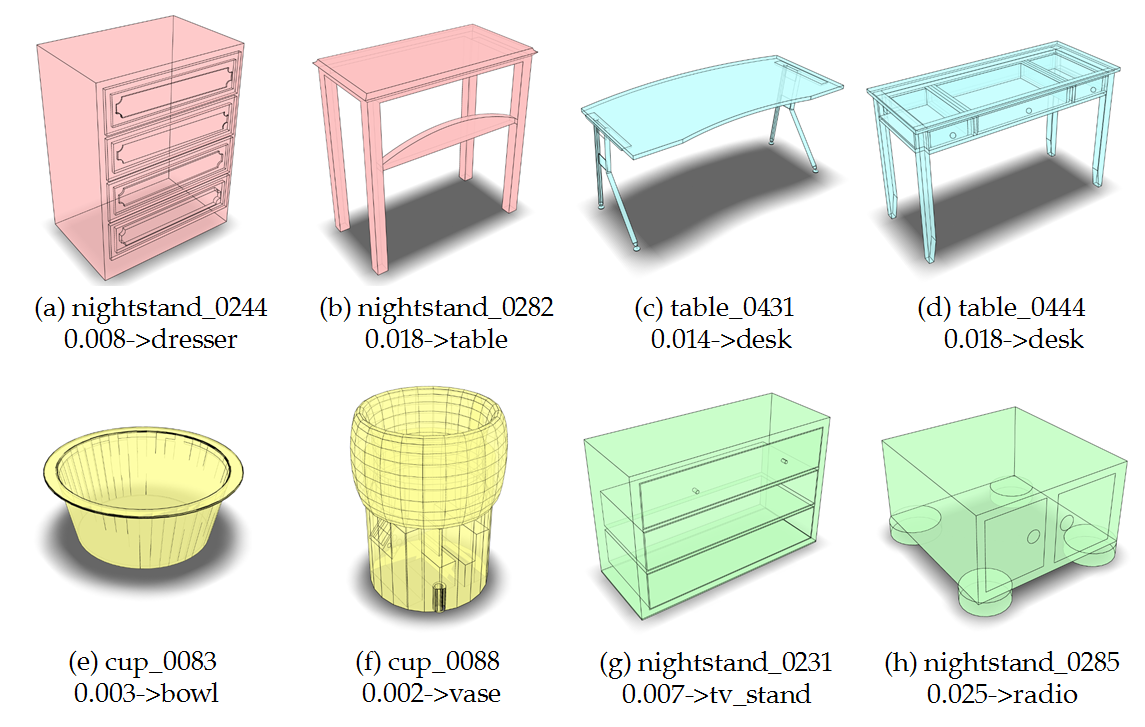}}
\caption{{\color{black}First row shows the examples of misclassified shapes of $nightstand$ and $table$ on ModelNet10. Second row shows the examples of misclassified shapes of $cup$ and $nightstand$ on ModelNet40. Shape name, probability to its own category and misclassified category are denoted below the shape.}}
\label{fig:misclassrevise}
\end{figure}}

\iffalse	

\begin{figure}[h]
\centering
{\includegraphics[width = .45\textwidth]{misclass.png}}
\caption{{\color{red}\sout{First row shows the examples of misclassified shapes of $nightstand$ and $dresser$ on ModelNet10. Second row shows the examples of misclassified shapes of $nightstand$ and $cup$ on ModelNet40. Shape name, probability to its own category and wrong category are denoted below the shape.}} }
\label{fig:misclass}
\end{figure}

 {\color{red}\sout{On ModelNet40, $nightstand$ is still the worst category and the accuracy of $nightstand$ is reduced to $64.0\%$. $Cup$ is the second worst category and the accuracy of $cup$ is $75.0\%$. Total $31$ shapes of $nightstand$ are misclassified. Among them, $16$ shapes are misclassified as $dresser$ and the others are spread to $8$ categories including $bookshelf$, $chair$, $desk$, etc. The testing set of ModelNet40 only contains $20$ $cups$. $5$ shapes of $cup$ are misclassified. One of them is misclassified as $bowl$, another is misclassified as $flowerpot$, and the other $3$ $cups$ are misclassified as $vase$. Two misclassified shapes of $nightstand$ and $cup$ with their category probability are shown in the second row of Fig.~\ref{fig:misclassrevise}.}}
\fi

 {\color{black}On ModelNet40, the classification accuracy of $cup$ is $60.0\%$ and is the worst among all categories. $nightstand$ gets $72.1\%$ accuracy and holds the second worst position. Total $8$ $cups$ among 20 are misclassified. One each of them is misclassified as $bottle$, $bowl$ and $wardrobe$, and the other $5$ $cups$ are misclassified as $vase$. The testing set of ModelNet40 contains $86$ $nightstands$ and $24$ are misclassified. Among them, $15$ shapes are misclassified as $dresser$ and the other $9$ $nightstands$ are spread to $5$ categories including $bookshelf$, $glassbox$, $radio$, $table$, $tvstand$. Two misclassified shapes of $cup$ and $nightstand$ are shown in the second row of Fig.~\ref{fig:misclassrevise}.}

According to our intuition, the shapes shown in Fig.~\ref{fig:misclassrevise} are hard to be correctly classified by only using the visual representations.
%{\color{red}\sout{It could be said that such task is also difficult for human beings. }}
At least partly, the mistake dues to a shape may has diversified functions. We suspect that more structure information in 3D space is needed to be fully exploited to meet such challenges.

\section{Conclusion}

We have presented VERAM, a recurrent attention model capable of actively selecting a sequence of views for highly accurate 3D shape classification. To address the problem commonly found in existing RNN-based attention models, i.e., the unbalanced training of the subnetworks, we propose to
%{\color{red}\sout{The classification subnetwork is easily overfitted while the view estimation one is usually poorly trained.}}
1) enhance the information flow of gradient backpropagation for the view estimation subnetwork, 2) design a highly informative reward function for the reinforcement training of view estimation, and 3) formulate a novel loss function that explicitly circumvents view duplication.

When being applied to real scenarios, VERAM has several limitations which may
spark future research:

\textit{Shape alignment.} %As \cite{johnse2016a, kanezakia2018a}, VERAM assumes all shapes in each category are well aligned.
The experiment in subsection 4.4 shows that the result is sensitive to alignment and this issue is also reported in\cite{kanezakia2018a}.
%However, the visual attention mechanism of human beings is not subject to this restriction.
 To address this problem, the basic strategy is by deepening the network and augmenting the training data to force the network to cover different viewpoint variants. Inspired by the approaches of learning transform parameters as \cite{Jaderbergm2015}, we think the more promising approach is to merge the alignment network into VERAM.

\textit{Time steps.}
VERAM uses the fixed time steps for predication.  %the accuracy is very close to the best performance when it is set to 6, as shown in Fig.~\ref{fig:numstepsrevise}.
%A more sophisticated visual attention model allows the network to learn to stop taking observation once it has enough information. For this purpose,%
To make the model capable of stopping observation once it has enough information, the reward of MV-RNN \cite{xuk2016a} contains the information gain and it terminates the process when entropy is less than the threshold, while \cite{yeungs2016a} designed a subnetwork to learn a binary prediction indicator. Both of them provide the insight to extend our model with varying instead of fixed time steps.

\textit{Inaccessible views and occlusion.}
%VERAM assumes all views are accessible and no any occupation.
In real scanning scenario, there are cases where some predicted views are physically inaccessible. Varying time steps is necessary and the proposed method also needs to be extended to learn the relevance of each view to the task as TAGM \cite{peiw2018}. Occlusion also has severe adverse effect since the feature representation of each view is by convoluting the entire image. The key to this problem is to encode the part-based feature as in \cite{xuk2016a} to spot informative visible parts of the partially occluded 3d shape.% and accurately discriminate shape category based on those parts.

\textit{Moving cost and views passed through.}
%The proposed method leaves out the cost of moving scanner, which also
The cost of moving scanner should be considered carefully in real scanning scenario. One feasible approach is to model the cost of moving scanner as the circle distance between adjacent selected views and add the cost to the reward for reinforce learning. Another issue is VERAM omits the continuous images obtained when moving the scanner from the current selected viewpoint to the next one. Although these views are not the best for next observation, but they also have the potential to help the classification to be more efficient.

\textit{Visual feature encoding.}
The pre-trained deep CNN is adopted by VERAM to extract visual features. For efficiency reason, such deep network is hard to fine-tuned with the view estimation subnetwork simultaneously, although what and where to observe are coupled for each other. It seems that visual attention model should exploit a more efficient network to learning what to observe.

\ifCLASSOPTIONcompsoc
  \section*{Acknowledgments}
\else
  \section*{Acknowledgment}
\fi
 %{\color{blue}We thank the anonymous reviewers for their valuable comments and suggestions.}
 This work was supported in part by the National Natural Science Foundation of China (61373135, 61672299, 61702281, 61532003, 61572507 and 61622212), and the Postdoctoral Science Foundation of Jiangsu Province of China (No.1701046A).

\ifCLASSOPTIONcaptionsoff
  \newpage
\fi



%\bibliographystyle{IEEEtran}

\begin{IEEEbiography}[{\includegraphics[width=1in,height=1.25in,clip,keepaspectratio]{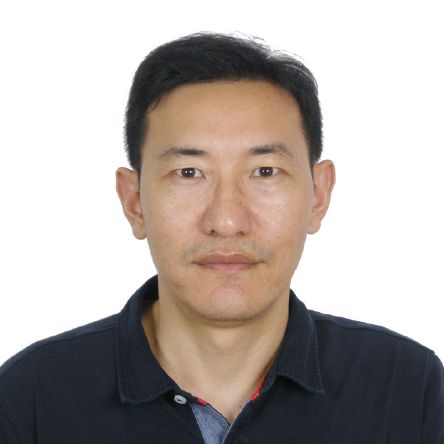}}]{Songle Chen}  received the Ph.D. degree from the Department of Computer Science and Technology of Nanjing University in 2015. Now he is a lecturer in Nanjing University of Posts and Telecommunications and a researcher of Jiangsu High Technology Research Key Laboratory for Wireless Sensor Networks. His research interests include Geometric Modeling, Realistic Rendering, Computer Animation, Deep Learning etc.
\end{IEEEbiography}

\begin{IEEEbiography}[{\includegraphics[width=1in,height=1.25in,clip,keepaspectratio]{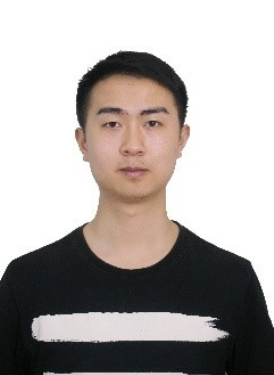}}]{Lintao Zheng} received his BS degree in applied mathematics from Xi¡¯an Jiaotong University and MS degree in computer science from the National University of Defense Technology in 2013 and 2016, respectively. He is pursuing a doctorate in computer science at the National University of Defense Technology. His research interests mainly include computer graphics, deep learning, robot vision. He is a member of the ACM.
\end{IEEEbiography}

\begin{IEEEbiography}[{\includegraphics[width=1in,height=1.25in,clip,keepaspectratio]{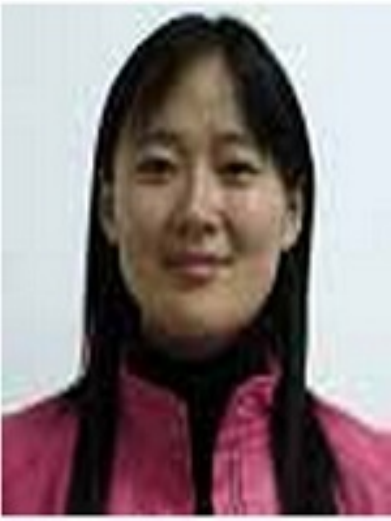}}]{Yan Zhang}received the Ph.D. degree from Jilin University in 2003 and finished his Post-doctoral researches in Nanjing University in 2006. She is now an associate professor of the Department of Computer Science and Technology in Nanjing University. Her research interests include Computer graphics, Image and Video Process and Computer Aided Design. Her research work can be found in her personal website: http://cs.nju.edu.cn/zhangyan.
\end{IEEEbiography}

\begin{IEEEbiography}[{\includegraphics[width=1in,height=1.25in,clip,keepaspectratio]{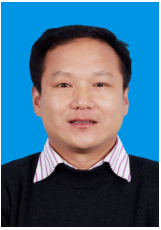}}]{Zhixin Sun} is the professor and dean of School and Institute of Modern Posts, Nanjing University of Posts and Telecommunications. He received his Ph.D.degree in Nanjing University of Aeronautics and Astronautics, China in 1998 and worked as a post doctor in Seoul National University, South Korea between 2001 and 2002. His research area includes Computer Aided Design, computer vision, information security, computer networks, etc.
\end{IEEEbiography}

\begin{IEEEbiography}[{\includegraphics[width=1in,height=1.25in,clip,keepaspectratio]{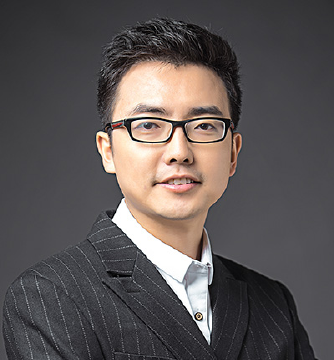}}]{Kai Xu}is an Associate Professor at the School of Computer, National University of Defense Technology, where he received his Ph.D. in 2011. He conducted visiting research at Simon Fraser University during 2008-2010, and Princeton University during 2017-2018. His research interests include geometry processing and geometric modeling, especially on data-driven approaches to the problems in those directions, as well as 3D-geometry-based computer vision. He has published over 60 research papers, including 21 SIGGRAPH/TOG papers. He organized two SIGGRAPH Asia courses and one Eurographics STAR tutorial. He is currently serving on the editorial board of Computer Graphics Forum, Computers \& Graphics, and The Visual Computer. He also served as paper co-chair of CAD/Graphics 2017 and ICVRV 2017, as well as PC member for several prestigious conferences including SIGGRAPH Asia, SGP, PG, GMP, etc. His research work can be found in his personal website: www.kevinkaixu.net.
\end{IEEEbiography}


%\vfill


\end{document}